\documentclass[pdflatex,sn-mathphys-num]{sn-jnl}

\usepackage{graphicx}%
\usepackage{multirow}%
\usepackage{amsmath,amssymb,amsfonts}%
\usepackage{amsthm}%
\usepackage{mathrsfs}%
\usepackage[title]{appendix}%
\usepackage{xcolor}%
\usepackage{textcomp}%
\usepackage{manyfoot}%
\usepackage{booktabs}%
\usepackage{algorithm}%
\usepackage{algorithmicx}%
\usepackage{algpseudocode}%
\usepackage{listings}%

\usepackage{geometry}
\theoremstyle{thmstyleone}%
\theoremstyle{thmstyletwo}%

\theoremstyle{thmstylethree}%

\begin{document}

\title[Article Title]{Earth Embeddings Reveal Diverse Urban Signals from Space}

\author[1]{\fnm{Wenjing} \sur{Gong}}
\equalcont{These authors contributed equally to this work.}
\author[2]{\fnm{Udbhav} \sur{Srivastava}}
\equalcont{These authors contributed equally to this work.}

\author[1]{\fnm{Yuchen} \sur{Wang}}
\author[3]{\fnm{Yuhao} \sur{Jia}}
\author[3]{\fnm{Qifan} \sur{Wu}}
\author[1]{\fnm{Weishan} \sur{Bai}}
\author[1]{\fnm{Yifan} \sur{Yang}}

\author*[3]{\fnm{Xiao} \sur{Huang}}\email{xiao.huang2@emory.edu}
\author*[1,2]{\fnm{Xinyue} \sur{Ye}}\email{xye10@ua.edu}

\affil[1]{\orgname{Texas A\&M University}, \orgaddress{\city{College Station}, \state{TX}, \country{USA}}}
\affil[2]{\orgname{The University of Alabama}, \orgaddress{\city{Tuscaloosa}, \state{AL}, \country{USA}}}
\affil[3]{\orgname{Emory University}, \orgaddress{\city{Atlanta}, \state{GA}, \country{USA}}}

\abstract{Conventional urban indicators derived from censuses, surveys, and administrative records are often costly, spatially inconsistent, and slow to update. Recent geospatial foundation models enable the generation of Earth embeddings, compact representations of satellite imagery that are transferable across downstream tasks, but their utility for neighborhood-scale urban monitoring remains unclear. Here, we systematically benchmark three Earth embedding families, AlphaEarth, Prithvi, and Clay, for urban signal prediction across six major U.S. metropolitan areas from 2020 to 2023. Using a unified supervised-learning framework, we predict 14 neighborhood-level indicators spanning crime, income, health, and travel behavior, and evaluate performance under four complementary settings: global, city-wise, year-wise, and city–year. Results show that Earth embeddings capture substantial urban variation, with the highest predictive skill for outcomes more directly tied to built-environment structure, including chronic health burdens and dominant commuting modes. By contrast, indicators more strongly shaped by fine-scale behavior and local policy, such as cycling, remain difficult to infer. Predictive performance varies markedly across cities but is comparatively stable across years, indicating strong spatial heterogeneity alongside temporal robustness. Exploratory analysis further suggests that cross-city variation in predictive performance is associated with urban form indicators, including population density, employment and household entropy, and walkability, in task-specific ways. Controlled dimensionality experiments show that representation efficiency is critical: compact 64-dimensional AlphaEarth embeddings remain more informative than 64-dimensional reductions of higher-dimensional Prithvi and Clay embeddings. This study establishes a benchmark for evaluating Earth embeddings in urban remote sensing and demonstrates their potential as scalable, low-cost features for neighborhood-scale urban monitoring aligned with the SDGs in data-scarce settings.}

\maketitle

\section{Introduction}\label{sec1}
Rapid global urbanization has catalyzed a complex suite of challenges, ranging from persistent poverty and health inequities to elevated crime rates and strained transportation systems\cite{Hidden,World,Worlda}. Addressing these multifaceted issues is central to the United Nations Sustainable Development Goals (SDGs), particularly Goals 1 (No Poverty), 3 (Good Health and Well-being), 11 (Sustainable Cities and Communities), and 13 (Climate Action)\cite{United}. As cities continue to grow and diversify, achieving these targets requires urban governance grounded in timely, comparable data and in the ability to monitor neighborhood-scale urban signals with high precision and continuity\cite{Unsdg,IAEGSDGs}.

The data infrastructures traditionally used to monitor these dimensions are increasingly ill-equipped to keep pace with rapid urban change. Conventional metrics, predicated on decennial censuses\cite{BureauDecennial}, periodic household surveys\cite{BureauAmerican}, and administrative records\cite{BureauAdministrative}, are frequently plagued by significant temporal lags, prohibitive collection costs, and fragmentation across disparate regions. Such data often fail to capture the near-real-time dynamics of urban vitality, leaving policymakers with an incomplete or outdated picture. Furthermore, while emerging alternative proxies such as street-view imagery have provided a rich lens into the local built environment\cite{Fan2023Urbana,Biljecki2021Street}, they remain constrained by uneven spatial coverage\cite{Fan2025Coverage}, stringent privacy regulations\cite{Frome2009Largescale}, and the substantial computational overhead required to process ground-level visual semantics at metropolitan or national scales\cite{Gebru2017Using}. Consequently, a critical information gap persists around the social distribution of urban progress, hindering our ability to track who benefits and who is left behind in cities in a consistent and equitable way.

The proliferation of Earth Observation (EO) data, encompassing nighttime lights products\cite{Roman2018NASAs}, the long-running Landsat program\cite{Wulder2022Fifty}, and the Sentinel missions\cite{Drusch2012Sentinel2}, offers a transformative opportunity for continuous, large-scale urban monitoring. For much of the past few decades, leveraging EO data for urban proxy estimation has relied on labor-intensive feature engineering, where researchers manually designed spectral indices, textural measures, or hand-crafted spatial statistics tailored to specific tasks, regions, and sensors\cite{Mellander2015NightTime,Stark2020SatelliteBased,Engstrom2022Poverty}. Such workflows are difficult to scale, sensitive to domain shifts, and offer limited transferability across indicators or cities\cite{Tuia2016Domain}. Recently, the field has undergone a paradigm shift toward self-supervised learning and the development of geospatial foundation models\cite{Xiao2025Foundation}. Cutting-edge models such as AlphaEarth Foundations\cite{Brown2025AlphaEarth}, Prithvi-EO-2.0\cite{Szwarcman2025PrithviEO20}, and Clay\cite{Clay} are designed to ingest massive volumes of unlabeled remote sensing imagery to generate Earth embeddings—universal latent vectors that encapsulate complex spatial textures, land-cover patterns, and high-level semantic information. These embeddings serve as widely applicable inputs for diverse downstream applications, moving beyond task-specific architectures to provide a unified representation of the Earth’s surface that can, in principle, be reused across tasks, locations, and time periods\cite{Klemmer2025Earth}.

These Earth embeddings have already demonstrated strong performance on physical sensing tasks. Prithvi-EO models, pretrained on multi-temporal Harmonized Landsat–Sentinel-2 imagery, reach competitive performance on land-cover\cite{Szwarcman2025PrithviEO20} and rice mapping\cite{Fang2026Generating}, in fine-tuned variants such as Prithvi-100M-sen1floods11, are used for flood-extent segmentation\cite{Introduction}. Clay, a ViT-based masked autoencoder trained on Sentinel-1/2 and DEM, outperforms other geospatial foundation models on land-cover segmentation\cite{Wiratama2025Comparative}. Additionally, AlphaEarth Foundations provides globally consistent 10 m embeddings that support high-accuracy environmental and agricultural mapping, including air pollution prediction\cite{Alvarez2025Machine}, flash flood forecasting\cite{Ashfaq2025TheoryGuided}, and crop yield estimation\cite{Ma2025Harvesting}. The success of these physical tasks is fundamentally rooted in the direct spectral signatures of biophysical phenomena. In contrast, urban indicators such as crime, poverty, health disparities, and travel behavior are more latent outcomes, emerging from the complex interplay between built-environment morphology and human mobility\cite{Smith2022Climate,Hall2023review,Gong2026Revealing}. Whether the universal embeddings learned from predominantly physical geographic features can successfully bridge this domain gap to capture subtle, human-centric signals remains a critical uncertainty in the field. To move beyond fragmented evidence and toward a systematic understanding of this new paradigm, our study seeks to address three pivotal research questions. First, to what extent can universal representations trained predominantly on physical geographic features capture neighborhood-level urban signals? Second, how robust are these embeddings across space and time—that is, how consistently do they perform across different urban contexts and across evolving temporal windows? Third, what trade-offs arise between alternative embedding architectures when they are applied to the multi-dimensional nature of urban life?

To this end, we evaluate three state-of-the-art Earth embeddings, namely AlphaEarth, Prithvi-EO-2.0-300M (Prithvi), and Clay-v1.5 (Clay), for neighborhood-level urban signal prediction across six major U.S. Metropolitan Statistical Areas (MSAs) from 2020 to 2023. Our analysis spans four urban dimensions, income, health, crime, and travel, aligned with SDGs 1, 3, 11, and 13 (see Supplementary Table 1 for details), and uses a unified supervised learning pipeline across four complementary evaluation settings: global, city-wise, year-wise, and city–year. This design enables us to assess both overall predictive skill and its sensitivity to spatial and temporal variation. We show that Earth embeddings recover substantial neighborhood-level variation, with the strongest performance for outcomes that are more tightly coupled to built-environment structure, including chronic health burdens and dominant commuting modes, whereas indicators such as cycling remain much harder to infer. We further find pronounced cross-city heterogeneity in predictive performance, while results remain comparatively stable across years, and exploratory analysis suggests that this variation is associated, in task-specific ways, with differences in urban form. Across most tasks and settings, AlphaEarth delivers the most reliable performance, and controlled dimensionality experiments further indicate that its compact 64-dimensional representation is more information-dense than compressed variants of higher-dimensional Prithvi and Clay embeddings.

This study presents the first systematic benchmark linking Earth embeddings to neighborhood-scale urban signal prediction across multiple cities, years, and outcome domains. By revealing performance trade-offs across tasks, spatial contexts, and embedding families, it offers a foundational roadmap for researchers and urban planners to deploy these embeddings in human-centered urban analytics. Furthermore, this scalable workflow demonstrates the practical potential of Earth embeddings as high-frequency proxies for urban monitoring, particularly in data-scarce regions where traditional survey methods are unavailable. Ultimately, this work offers a new lens, as seen from space, through which to observe the social fabric of cities and advance progress toward global sustainable development.

\section{Results}\label{sec2}

\begin{figure}[h]
\centering
\includegraphics[width=0.9\textwidth]{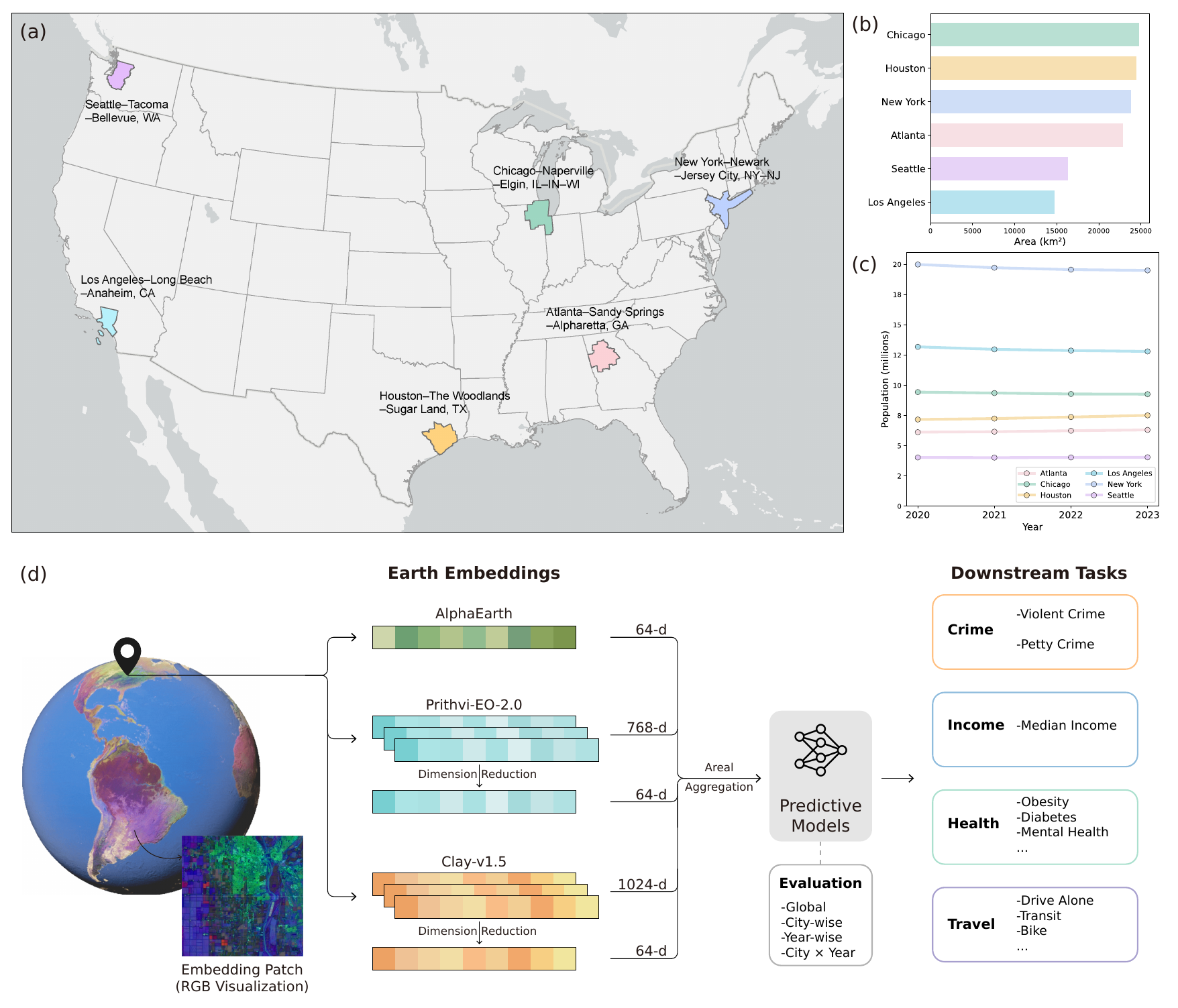}
\caption{Study area and research framework. (a) The selected six MSAs: Atlanta–Sandy Springs–Alpharetta, GA (Atlanta), Chicago–Naperville–Elgin, IL–IN–WI (Chicago), Houston–The Woodlands–Sugar Land, TX (Houston), Los Angeles–Long Beach–Anaheim, CA (Los Angeles), New York–Newark–Jersey City, NY–NJ (New York),  and Seattle–Tacoma–Bellevue, WA (Seattle). (b) Land area (km²) of the six MSAs. (c) Total population from 2020 to 2023 for each MSA. (d) Conceptual framework of Earth-embedding-based urban signal prediction.}\label{fig1}
\end{figure}

In this study, we collected three widely used Earth embeddings (AlphaEarth, Prithvi, and Clay) for six U.S. MSAs from 2020 to 2023, spanning cities that differ markedly in population size and geographic setting (Fig. \ref{fig1}a–c). As shown in Fig. \ref{fig1}d, we evaluated these representations for predicting 14 target urban indicators grouped into four domains: crime, income, health, and travel. For example, we used obesity, diabetes, no leisure-time physical activity (inactivity), poor mental health, poor physical health, and cancer prevalence to represent health, and commuting-mode shares, such as public transit, car, walking, and cycling to represent travel (see Supplementary Figs.~1--4 for details). Earth embeddings were spatially aggregated to U.S. census block groups (CBGs), except for health indicators, which are only available at the census tract (CT) level and were therefore analyzed at that coarser spatial resolution. Maps of Earth embeddings aggregated to CBGs are presented in Supplementary Figs. 5–7. 

Across all analyses, we applied a unified supervised-learning pipeline at four complementary evaluation granularities: global, city-wise, year-wise, and city–year. For each configuration, the corresponding spatial units were randomly divided into a training set (80\%) and a held-out test set (20\%). We trained four models, Ordinary Least Squares (OLS) Regression, Random Forest, XGBoost, and LightGBM, and used five-fold cross-validation within the training set to select the model with the highest mean cross-validated R². Unless otherwise noted, we report test R² from LightGBM, which achieved the highest cross-validated performance for most prediction tasks, and present the full model comparison results under the global setting in Supplementary Figs.~8–11.

\subsection{Earth embeddings capture neighborhood-level urban variation}\label{subsec2.1}

\begin{figure}[h]
\centering
\includegraphics[width=0.9\textwidth]{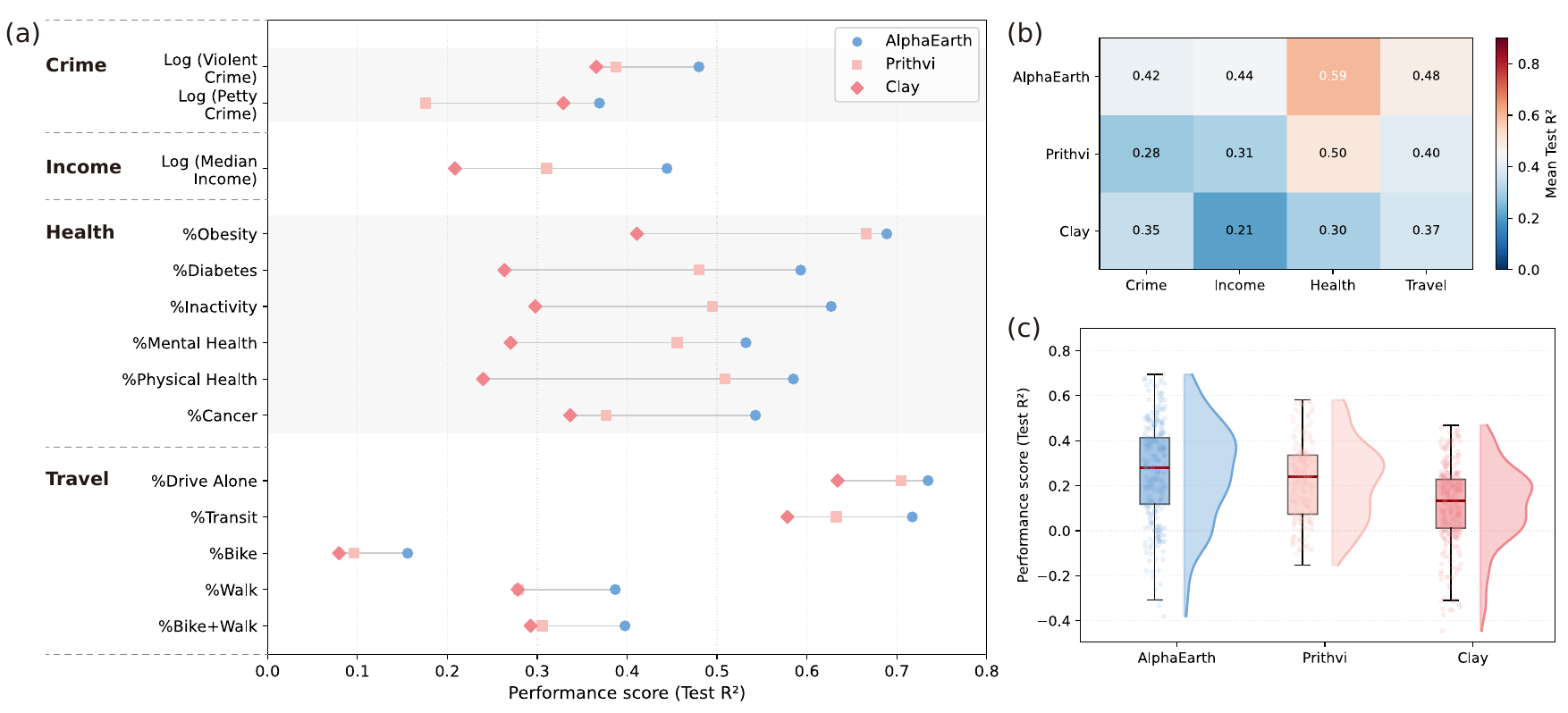}
\caption{Benchmarking Earth embeddings for predicting urban signals. (a) Global predictive performance (test R²) of three Earth embeddings (AlphaEarth, Prithvi, and Clay) across 14 urban indicators spanning crime, income, health, and travel behavior for six US MSAs over 2020–2023 (see Supplementary Table 2 for details). (b) Global mean R² by thematic domain, obtained by averaging across indicators, cities, and years, summarizing domain-level differences in model skill. (c) Distribution of city–year experiment test R² values across all indicators, showing the variability and upper-tail behavior of each embedding.}\label{fig2}
\end{figure}

We first evaluated the three Earth embeddings at a global scale across 14 urban indicators spanning crime, income, health, and travel behaviour for six MSAs over 2020–2023 (Fig. \ref{fig2}a). Overall, performance varied markedly by outcome and embedding families. AlphaEarth achieved the highest and most consistent predictive skill, with Prithvi usually ranking second and Clay delivering slightly lower scores for most variables.  Health indicators are generally easiest to predict, with \%Obesity remaining high for both AlphaEarth and Prithvi (0.69 and 0.67, respectively) and \%Inactivity following a similar pattern (0.63 vs. 0.50). In travel, predictability varies substantially across commuting modes: driving and transit are captured best—\%Drive Alone reaches 0.74 (AlphaEarth) and 0.71 (Prithvi), while Clay still performs well at 0.63; \%Transit shows the same ordering (0.72, 0.63, and 0.58). By contrast, active-transport remains difficult, with \%Bike staying low across all three embeddings (0.16, 0.10, and 0.07). Finally, income and crime show more mixed predictability. Log-transformed median household income is only moderately captured (AlphaEarth, 0.44; Prithvi, 0.31; Clay, 0.21). Within crime, log violent-crime rate is more predictable than log petty-crime rate overall (violent, 0.48/0.39/0.37 versus petty, 0.37/0.18/0.33 for AlphaEarth/Prithvi/Clay).

Aggregating indicators into four thematic groups confirms that AlphaEarth embeddings are most informative for capturing urban profiles across all domains (Fig. \ref{fig2}b). AlphaEarth attains the best mean R² for health (0.59) and shows strong performance for crime (0.42), income (0.44), and travel (0.48). Prithvi achieves mean R² values ranging from 0.28 to 0.50 across groups, while Clay generally trails, particularly for income (mean R² = 0.21), despite moderate performance for crime (0.35), health (0.30), and travel (0.37). The results of the city–year experiment across all indicators further show that these differences are robust rather than driven by a few favourable cases (Fig. \ref{fig2}c). AlphaEarth exhibits the highest median R² and a heavier upper tail, indicating more frequent instances of strong predictive performance, whereas Clay shows lower central tendencies and a greater share of near-zero or negative R² values.

\subsection{Cross-city heterogeneity in urban signal predictability}\label{subsec2.2}
\begin{figure}[h]
\centering
\includegraphics[width=0.9\textwidth]{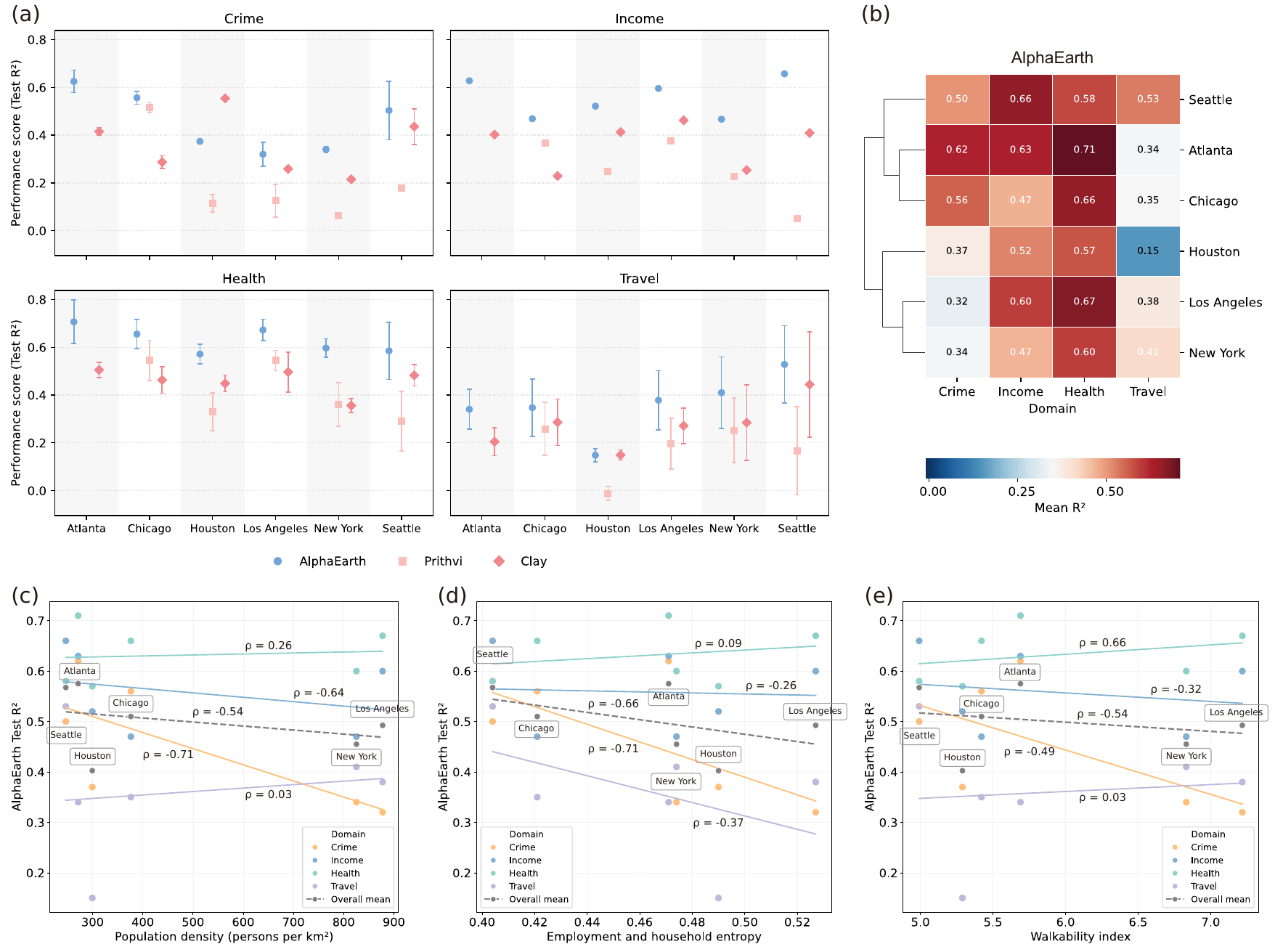}
\caption{Cross-city heterogeneity in the predictability of urban signals. (a) City-wise predictive performance (test R²) of three Earth embeddings (AlphaEarth, Prithvi, and Clay) across four domains for six US MSAs over 2020–2023. (b) Hierarchical clustering of MSAs based on AlphaEarth’s mean R² across domains. (c–e) Exploratory relationships between urban form indicators and domain-specific AlphaEarth predictive performance: (c) population density, (d) employment and household entropy, and (e) walkability index. Colored lines show simple domain-specific trend lines, and $\rho$ indicates Spearman’s rank correlation.}\label{fig3}
\end{figure}

We next examined how the predictive skill of the Earth embeddings varies across MSAs for each urban indicator over the 2020--2023 period (Supplementary Fig.~12), and subsequently aggregated these multi-year indicator-level scores to the thematic domain level (Fig. \ref{fig3}a). Across domains, the results reveal clear between-city contrasts in how well Earth embeddings predict urban outcomes. Atlanta, Seattle, Chicago, and Los Angeles generally appear as “easier” cities: they sustain relatively high R$^2$ for income and health. For instance, AlphaEarth reaches about 0.63/0.70 (income/health) in Atlanta, 0.66/0.59 in Seattle, 0.47/0.66 in Chicago, and 0.60/0.67 in Los Angeles; and in several cases, they also achieve moderate performance for crime or travel (e.g., Atlanta crime R$^2$ $\sim$ 0.62, Seattle travel R$^2$ $\sim$ 0.53). By contrast, Houston tends to be systematically harder to predict, particularly for travel, where all three models cluster at low R$^2$ (AlphaEarth $\sim$ 0.15, Clay $\sim$ 0.15) and Prithvi even attains near-zero to slightly negative values ($\sim -0.01$). Clay occasionally matches or exceeds AlphaEarth's accuracy, for example, in predicting crime in Houston (R$^2$ $\sim$ 0.55 for Clay vs.\ $\sim$ 0.37 for AlphaEarth), highlighting that different Earth embeddings capture partially complementary aspects of urban profiles.

We synthesised these patterns by clustering cities based on AlphaEarth's mean R$^2$ across domains (Fig. \ref{fig3}b), as AlphaEarth was the strongest model in most domains and cities. Hierarchical clustering on the mean R$^2$ for crime, income, health, and travel separates the six MSAs into two main groups. Seattle, Atlanta, and Chicago form a cluster characterised by consistently high predictability across domains (e.g., income/health mostly $\sim$0.5--0.7), whereas Los Angeles, New York, and Houston form a second group in which these MSAs retain high income and health performance (again typically $\sim$0.5--0.7) but much poorer predictability for crime (roughly $\sim$0.3--0.4) and, in some cases, travel (most notably Houston travel R$^2$ $\sim$ 0.15). This structure indicates that Earth embeddings transfer across heterogeneous urban contexts, but predictive effectiveness remains strongly modulated by city-specific conditions, producing distinct clusters of cities that are comparatively easier or harder to predict.

To further probe the city-specific conditions underlying this cross-city heterogeneity for AlphaEarth, we conducted an exploratory comparison using three urban form indicators: population density, employment and household entropy, and walkability (Fig. \ref{fig3}c--e). For each urban form indicator, we fitted domain-specific trend lines and calculated Spearman’s rank correlations. Crime prediction showed the clearest and most consistent negative relationship with urban form, with Spearman's $\rho$ equal to $-0.71$ for both population density and jobs--housing entropy, and remaining negative for walkability ($\rho = -0.49$). Income prediction also showed a relatively strong negative association with population density ($\rho = -0.64$). By contrast, health prediction showed the strongest positive tendency for walkability ($\rho = 0.66$), whereas travel exhibited a moderate negative pattern with employment and household entropy ($\rho = -0.37$). These exploratory findings suggest that urban contextual characteristics relate to predictive performance in complex, task-specific ways that deserve further investigation.

\subsection{Temporal robustness of urban signal prediction from 2020 to 2023}\label{subsec2.3}

We further explored temporal dynamics in model skill by computing, for each year from 2020 to 2023, the mean R$^2$ for each indicator in the pooled multi-MSA experiments and aggregating these values at the thematic domain level. As shown in Supplementary Fig.~13, the R$^2$ values of indicator-level results are largely stable over time, with only modest year-to-year fluctuations (typically within $\sim$0.02--0.06 for most indicators). Earth embeddings perform best on some health and travel indicators: for example, \%Drive Alone remains high across 2020--2023 (about 0.68--0.74 for AlphaEarth, 0.52--0.75 for Prithvi, and 0.60--0.64 for Clay), and \%Obesity also stays consistently strong (roughly 0.62--0.69 for AlphaEarth, 0.62--0.66 for Prithvi, and 0.28--0.35 for Clay). In contrast, the percentage of commuters traveling by bicycle remains one of the least predictable indicators throughout 2020--2023, with R$^2$ values staying near 0.1 for AlphaEarth and Prithvi, and even lower for Clay (0.03).

\begin{figure}[h]
\centering
\includegraphics[width=0.9\textwidth]{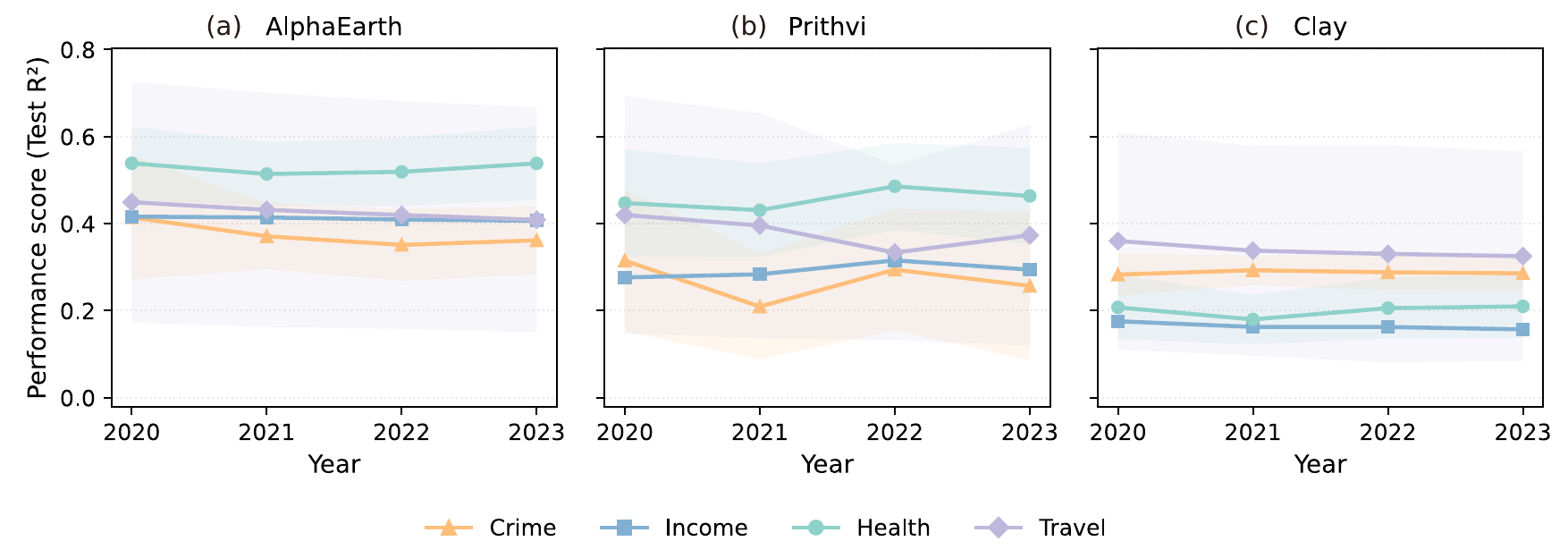}
\caption{Consistency of urban signal prediction across years (2020–2023). Annual predictive performance (test R²) for four urban domains (crime, income, health, and travel) from 2020 to 2023, aggregated across all indicators and MSAs for (a) AlphaEarth, (b) Prithvi, and (c) Clay. Lines show the domain-wise mean R² in each year, and shaded bands indicate variability across indicators.}\label{fig4}
\end{figure}

At the thematic domain level, this temporal stability becomes even more evident: across all three Earth embeddings, domain-level R$^2$ values remain remarkably consistent from 2020 to 2023, indicating that the embeddings transfer robustly to urban prediction across multiple years. For AlphaEarth, health remains the most predictable domain (R$^2$ $\sim$ 0.51--0.54), followed by travel ($\sim$ 0.41--0.45) and income ($\sim$ 0.41--0.42), with crime consistently emerging as the hardest to capture ($\sim$ 0.35--0.41) (Fig. \ref{fig4}a). Prithvi exhibits a similar domain ordering at lower absolute levels and with larger inter-annual fluctuations, including a pronounced dip in crime predictability in 2021 ($\sim$ 0.21), followed by partial recovery and a peak in 2022 ($\sim$ 0.30) (Fig. \ref{fig4}b). Clay displays a distinct pattern with minimal temporal variation across domains, with travel as the most predictable outcome (R$^2$ $\sim$ 0.33--0.36), whereas income and health cluster at lower values ($\sim$ 0.16--0.21) (Fig. \ref{fig4}c).

\subsection{Information density and representation efficiency}\label{subsec2.4}

So far, AlphaEarth has outperformed Prithvi and Clay across most indicators and domains. However, these embeddings differ substantially in dimensionality: Prithvi and Clay produce much higher-dimensional embeddings (768-d vectors by default for Prithvi-EO-2.0-300M and 1024-d vectors for Clay v1.5) than AlphaEarth’s 64-d representation. This raises the possibility that part of their weaker performance reflects redundant or noisy dimensions rather than fundamentally poorer representations. To isolate the contribution of representation quality from mere vector size, we conducted a controlled experiment in which we reduced Prithvi and Clay embeddings to 64 dimensions using five standard dimensionality-reduction techniques (factor analysis, Isomap, kernel PCA, PCA, and random projection), matched to AlphaEarth’s size, and re-evaluated their performance (see Methods for details).

\begin{figure}[h]
\centering
\includegraphics[width=0.9\textwidth]{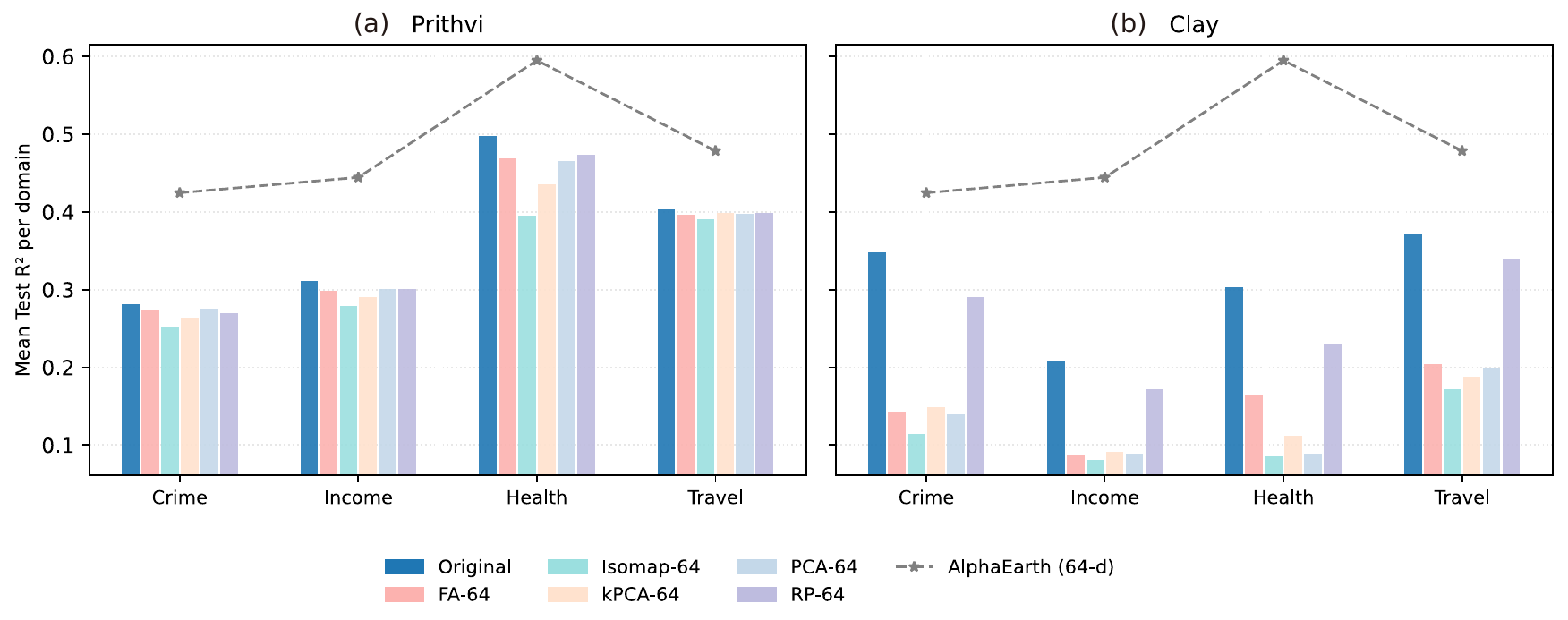}
\caption{Global information density and representation efficiency of Earth embeddings. Mean predictive performance (test R²) by domain (crime, income, health, and travel) for (a) Prithvi and (b) Clay using their original high-dimensional embeddings and five 64-d compressed variants (FA-64, Isomap-64, kPCA-64, PCA-64, and RP-64). Bars show domain-wise mean R² aggregated across all indicators, MSAs, and years. The dashed line with stars denotes the original 64-d AlphaEarth performance. }\label{fig5}
\end{figure}

The results benchmarked on a global scale show that compressing Prithvi and Clay to 64 dimensions never improves their performance; in all domains, the reduced Earth embeddings perform worse than their original high-dimensional counterparts, with the strongest degradation observed for Clay. For Prithvi, dimensionality reduction leads to modest but systematic declines in mean R² (Fig. \ref{fig5}a). For example, health decreases from about ~0.50 to roughly ~0.40–0.47, while crime drops slightly from ~0.28 to around ~0.25–0.27. Clay is markedly more sensitive to compression: reducing its embeddings to 64 dimensions causes pronounced drops, especially for crime and income, where performance can fall from around ~0.35 and ~0.21 to ~0.1-level values under several reduction methods (Fig. \ref{fig5}b). Notably, random projection (RP-64) consistently performs better than the other reduction methods for Clay. Across all settings, the 64-d AlphaEarth embeddings still achieve higher mean R² (roughly ~0.42–0.60 across domains) than any compressed variant of Clay or Prithvi, indicating that AlphaEarth provides more information-dense and representation-efficient embeddings. The indicator-level results can be found in Supplementary Tables 3 and 4.

\section{Discussions}\label{sec3}

A persistent gap in neighborhood-scale urban monitoring is the lack of timely and comparable data that can be acquired at low cost across cities and regions. Although Earth embeddings are widely used for physical mapping tasks, their application to neighborhood urban signal prediction has received limited attention, and cross-city and temporal transferability remains unresolved. This study systematically assesses the extent to which three widely used Earth embeddings encode socioeconomic signals in urban neighborhoods across six U.S. metropolitan areas. We find that current Earth embeddings can recover a substantial share of neighborhood-level urban variation, but their utility is strongly domain-dependent, modulated by city-specific context, and shaped by differences in representation efficiency across embedding families.

In this study, we found that Earth embeddings are most informative when outcomes are tightly coupled to the built environment and land-surface morphology that satellite imagery can directly encode. Health outcomes and dominant commuting modes exhibit relatively high predictability. This pattern implies that embeddings capture a stable suite of spatial signatures, such as density and accessibility gradients, impervious surface extent, vegetation and greenness structure, and infrastructure layouts, that are known to co-vary with chronic health risk burdens and with coarse travel choices\cite{Xu2025Predicting,Chen2024Deep,Yi2024SatelliteBased}. In contrast, indicators that depend more heavily on fine-grained behavioral preferences, institutional context, and unobserved social mechanisms are difficult to infer from Earth embeddings alone. Cycling share remains among the least predictable indicators across years and models, which mirrors the transportation literature showing that cycling uptake is shaped by determinants that are often micro-scale, policy-mediated, and perception-dependent, such as the presence and connectivity of bikeway networks, intersection treatments, perceived safety, traffic stress, and supportive packages of pro-cycling programs and restrictions on car use\cite{Pucher2010Infrastructure,Buehler2016Bikeway}. These determinants are often weakly expressed or inconsistently expressed at the spatial resolution and sensing modalities available to current embedding products. In this sense, Earth embeddings are best interpreted as structural priors for certain urban signal conditions, rather than as substitutes for ground-based measurement\cite{Engstrom2022Poverty,Jean2016Combining}.

The benchmark reveals a clear separation across embedding families, with AlphaEarth providing the most reliable signal for neighbourhood-scale urban prediction. Notably, this advantage is not a simple function of vector size: compressing Prithvi and Clay to a 64-d space does not improve performance and often degrades it, especially for Clay, suggesting that AlphaEarth may provide more information-dense and representation-efficient features. We also found that random projection consistently performs much better than the other reduction methods for Clay. A plausible explanation is that Clay embeddings, as patch-level explicit visual representations, may encode useful predictive information in a highly distributed manner rather than along a small set of dominant variance directions\cite{Clay,Klemmer2025Earth}. In that case, linear or nonlinear methods such as PCA, kPCA, and Isomap may discard or distort weaker but task-relevant components, whereas random projection may retain more globally useful structure under strong compression\cite{Dasgupta2003elementary}. More broadly, these results suggest that representation quality, as well as the alignment between pretraining signals and urban textures, matters more than raw embedding dimensionality. From an applied perspective, representation efficiency is especially consequential: compact embeddings reduce storage and computational overhead and are easier to integrate into practical urban analytics pipelines, consistent with a broader shift toward treating embeddings as scalable geospatial data layers\cite{Klemmer2025Earth,Blumenstiel2024MultiSpectral}.

Another major insight is that urban signal predictability from Earth embeddings varies systematically across cities. Several metropolitan areas (notably Atlanta, Seattle, and Chicago) are comparatively “easier” contexts, whereas Houston is consistently more challenging, most prominently for travel-related indicators, where all embeddings exhibit low skill. The resulting city clusters indicate that Earth embeddings can transfer across heterogeneous urban environments, but their effectiveness is conditioned on city-specific factors that determine how strongly socioeconomic gradients are expressed in land surface patterns and urban form\cite{Xu2025Predicting,Lakshminarayanan2017Simple}. To probe the urban conditions underlying this heterogeneity, our exploratory analysis of AlphaEarth suggests that urban form conditions predictive performance in task-specific ways rather than along a single gradient of city complexity. For crime and income, the negative associations with population density or functional mix suggest that, in more complex metropolitan environments, the spatial expression of neighborhood differences in these domains may align less well with the cues captured by Earth embeddings, contributing to lower predictive performance. The negative association between travel prediction and employment and household entropy further suggests that travel-related patterns may also become harder to predict in more functionally mixed metropolitan environments, where mobility outcomes are less tightly tied to visible urban form alone. By contrast, the positive tendency between walkability and health prediction suggests that health-related variation may be more predictable where neighborhood form more closely reflects lived environmental conditions. Beyond urban form, differences in administrative measurement systems across jurisdictions, such as variation in crime reporting standards and participation, may also contribute to cross-city differences in performance\cite{National}. Importantly, Clay occasionally matches or exceeds AlphaEarth in particular city–domain combinations (for example, certain crime predictions in Houston), suggesting that different embedding families may capture partially complementary urban signals. This complementarity motivates ensemble approaches or multi-embedding strategies that prioritize robustness under domain shift, especially in cities where any single embedding exhibits persistent weaknesses\cite{Zhou2023Domain}.

Despite pronounced spatial heterogeneity, the overall relationship between Earth embeddings and urban signal prediction is notably stable over time. From 2020 to 2023, performance within each domain shows only modest year-to-year fluctuation, and the relative domain ordering remains consistent for a given embedding family. This temporal robustness suggests that, over multi-year horizons, embeddings primarily encode slowly varying structural correlates of urban conditions rather than short-lived shocks. This is consistent with evidence that multi-temporal EO foundation models and self-supervised pretraining frameworks explicitly learn transferable representations from time series imagery, and that time-series satellite inputs can support longitudinal urban profile mapping\cite{Pettersson2023Time,CongSatMAEa}. Consequently, Earth embeddings appear well-suited for repeated, comparable monitoring, particularly when conventional survey-based indicators are sparse, inconsistent, or lagged. This suggests that embeddings can complement traditional data infrastructures without requiring frequent model redesign\cite{Klemmer2025Earth}, at least within the relatively short time window tested here and under consistent data products.

These results carry direct implications for urban sustainable development monitoring. First, Earth embeddings can function as scalable, low-cost features for high-frequency proxy estimation of neighborhood health burdens and dominant commuting patterns, enabling harmonized cross-city comparisons when traditional data infrastructures are incomplete or inconsistent. Second, the strong city dependence cautions against relying on a single global performance summary in isolation; operational deployments should incorporate city-specific calibration, uncertainty communication, and diagnostics for domain shift\cite{Al-Emadi2025Analysing,Ovadia2019Can}. Third, the evidence on representation efficiency indicates that compact embeddings can be highly competitive, lowering storage and compute barriers for large-area deployment without sacrificing predictive utility, provided the representation is information-dense and well aligned to urban textures.

Several limitations remain and point to clear directions for future work. First, although this study spans six major U.S. metropolitan areas, broader validation across a larger number of MSAs and across cities in other countries will be essential for assessing the global generalizability of Earth embeddings for urban signal prediction. Second, the selected outcomes were aggregated to administrative units (CBG/CT), introducing sensitivity to the modifiable areal unit problem (MAUP) and to boundary-specific aggregation choices\cite{Buzzelli2020Modifiable}. Third, outcomes sourced from the ACS were derived from 5-year rolling period estimates with substantial overlap across adjacent releases, which likely attenuates true year-to-year variation; therefore, our annual labels should be interpreted as end-year period estimates rather than independent single-year snapshots. Extending this evaluation to longer horizons and to periods with clearer built-environment change signals will be essential for understanding when temporal transfer breaks down and for identifying the conditions under which embeddings require updating or targeted fine-tuning. In addition, a particularly promising direction is multi-modal fusion that preserves cross-city portability while incorporating signals with stronger behavioral and institutional content, such as mobility traces, points-of-interest, administrative microdata, or micro-scale street-level features. Such fusion could help overcome persistent blind spots for outcomes like cycling and certain crime measures, where determinants are only weakly visible from overhead imagery.

\section{Methods}\label{sec4}

\subsection{Study site}\label{subsec4.1}
This study focuses on six MSAs in the United States: Atlanta–Sandy Springs–Alpharetta, GA (Atlanta), Chicago–Naperville–Elgin, IL–IN–WI (Chicago), Houston–The Woodlands–Sugar Land, TX (Houston), Los Angeles–Long Beach–Anaheim, CA (Los Angeles), New York–Newark–Jersey City, NY–NJ (New York),  and Seattle–Tacoma–Bellevue, WA (Seattle). These MSAs span the West Coast, Midwest, Northeast, and South, covering a wide range of climates, urban forms, and socioeconomic conditions (Fig. \ref{fig1}a–c).


\subsection{Earth embeddings}\label{subsec4.2}
We benchmarked three Earth embeddings that provide dense, fixed-length representations of satellite imagery: AlphaEarth, Prithvi-EO-2.0 (Prithvi), and Clay-v1.5 (Clay). All three models are trained in a self-supervised fashion on globally distributed EO data and expose generic feature embeddings that can be reused for downstream tasks. In this study, we extracted embeddings from 2020 to 2023 over the six selected MSAs. Pixel- or patch-level embeddings were reprojected, resampled, and aggregated to the target geographic units used for urban signal prediction.

Google DeepMind’s AlphaEarth Foundation (AEF) model is a geospatial foundation model with on the order of hundreds of millions of parameters that learns a unified latent representation of the Earth from multi-source EO data\cite{Brown2025AlphaEarth}. AEF is trained on a multi-year archive of optical imagery and ancillary geophysical variables sampled at millions of locations worldwide, using a spatio-temporal encoder–decoder architecture with an information bottleneck to capture continuous-time EO dynamics. In its “foundation” configuration, AEF produces 64-d embeddings at 10 m spatial resolution that summarise local land cover, morphology, and climate-related patterns. We used the publicly released 64-d surface embeddings from Google Earth Engine as fixed, off-the-shelf features without any additional fine-tuning.

Prithvi-EO-2.0 is a multi-temporal geospatial foundation model jointly released by NASA and IBM, which explicitly leverages transformer attention across both spatial and temporal dimensions\cite{Szwarcman2025PrithviEO20}. It is pre-trained on approximately a decade of medium-resolution optical imagery from the Harmonized Landsat–Sentinel-2 archive, enabling it to encode seasonal and inter-annual dynamics. To generate these embeddings, we retrieved multispectral scenes for six MSAs from 2020 to 2023 from the Harmonized Landsat Sentinel-2 (HLS) archive via the NASA EarthData API, specifically selecting the six bands required by the model architecture (Blue, Green, Red, NIR, SWIR1, and SWIR2). The imagery was then passed through the Prithvi Model to generate the embeddings. Prithvi produces high-dimensional feature vectors (768-d vectors by default) for each image patch; we used these original embeddings for our main benchmark and later considered compressed 64-d variants to compare representation efficiency with AlphaEarth.

Clay is an open-source EO foundation model trained with a masked autoencoding objective on globally distributed multi-sensor Earth observation data \citep{Clay}. It is designed to process inputs from different sensors, resolutions, and band configurations, and to produce high-dimensional embeddings that capture local geospatial characteristics. In this study, we used the pretrained Clay-v1.5 encoder without fine-tuning to generate static embeddings over the six MSAs from 2020 to 2023. Input imagery for Clay was sourced from the Sentinel-2 Level-2A archive via the Microsoft Planetary Computer STAC API, utilizing 10 optical bands at 10 m resolution. We processed the imagery into 256x256 pixel patches and extracted the 1024-d CLS token from the pretrained Clay-v1.5 encoder as the local feature embedding. As with Prithvi, we evaluated both the original Clay embeddings and 64-d compressed versions when analyzing information density.


\subsection{Target neighborhood-level urban signals}\label{subsec4.3}
We focused on 14 neighborhood-level urban signals organized into four thematic domains that are closely aligned with the United Nations’ SDGs: crime, income, health, and travel behavior. Summary statistics for all indicators used in this study are reported in Table \ref{tab1}.

Crime is represented by two indicators derived from police-recorded incidents available through each city’s open-data portal: the violent crime rate and the petty crime rate, each defined as the number of incidents per 10,000 residents within each CBG\cite{marquet2019short}. To reduce skewness and stabilize variance, both crime measures are log-transformed before modeling. Income is captured by median household income from the American Community Survey 5-year estimates (ACS 5-year)\cite{BureauAmericana}, also log-transformed to compress the long right tail typical of income distributions. Health conditions are characterized using six prevalence measures from the Centers for Disease Control and Prevention PLACES dataset: obesity, diabetes, no leisure-time physical activity (inactivity), poor mental health, poor physical health, and cancer\cite{Data}. These indicators are available at the census-tract level and are expressed as percentages of the adult population. Travel behavior is represented by five commuting-mode indicators from the ACS 5-year data: the shares of workers who drive alone, use transit, cycle, walk, and walk and cycle combined\cite{BureauAmericana}. Together, these four domains provide a compact yet diverse set of urban signal outcomes against which to benchmark the predictive value of Earth embeddings.

\begin{table}[h]
\centering
\caption{Summary statistics of urban signals for six MSAs from 2020 to 2023.}\label{tab1}
\begin{tabular}{llrrrrr}
\toprule
Theme & Indicator & count & mean & s.d. & min & max \\
\midrule
Crime & Log (Violent Crime) & 328,960 & 0.924 & 2.112 & 0.000 & 15.981 \\
      & Log (Petty Crime)   & 328,960 & 1.030 & 2.294 & 0.000 & 15.934 \\

Income & Log (Median Income) & 306,284 & 11.255 & 0.548 & 7.824 & 12.429 \\

Health & \%Obesity          & 287,850 & 32.045 & 7.444 & 10.400 & 64.400 \\
       & \%Diabetes         & 287,850 & 11.567 & 3.706 & 0.600 & 50.800 \\
       & \%Inactivity       & 287,850 & 25.033 & 8.273 & 6.900 & 75.500 \\
       & \%Mental Health    & 287,850 & 16.146 & 3.163 & 5.900 & 38.500 \\
       & \%Physical Health  & 287,850 & 12.391 & 3.540 & 2.300 & 36.400 \\
       & \%Cancer           & 287,850 & 6.546 & 2.300 & 0.500 & 24.700 \\

Travel & \%Drive Alone      & 331,783 & 67.621 & 22.230 & 0.000 & 100.000 \\
       & \%Transit          & 331,783 & 7.123  & 15.130 & 0.000 & 100.000 \\
       & \%Bike             & 331,783 & 0.552  & 2.181  & 0.000 & 100.000 \\
       & \%Walk             & 331,783 & 2.840  & 6.769  & 0.000 & 100.000 \\
       & \%Bike+Walk        & 331,783 & 3.392  & 7.381  & 0.000 & 100.000 \\
\bottomrule
\end{tabular}
\footnotesize{Note:} This table summarizes all urban indicators at the CBG level, except for indicators in the health theme, which are only available at the CT level.
\end{table}


\subsection{Predicting urban signals with Earth embeddings}\label{subsec4.4}
We implemented a unified supervised learning framework to evaluate the predictive utility of the Earth embeddings. For each task, we utilized four algorithms representing different levels of model complexity: (1) Ordinary Least Squares (OLS) Regression, providing a baseline for linear separability\cite{Zdaniuk2014Ordinary}; (2) Random Forest, a classic ensemble bagging technique\cite{Breiman2001Random}; (3) XGBoost, a scalable end-to-end tree boosting system\cite{Chen2016XGBoost}; and (4) LightGBM, a highly efficient gradient boosting framework that utilizes Gradient-based One-Side Sampling\cite{Ke2017LightGBM}. For indicators characterized by significant zero-inflation (e.g., crime rates), we employed a two-part Hurdle model approach, using a binary classifier to predict event occurrence followed by a regressor for the continuous magnitude\cite{Hu2011Zeroinflated}. We evaluated model performance under four complementary settings, namely global, city-wise, year-wise, and city–year, to assess predictive skill at different spatial and temporal granularities. For each setting, spatial units were randomly split into a training set (80\%) and a held-out test set (20\%). Model selection and tuning were conducted using five-fold cross-validation within the training set, and final performance was reported on the held-out test set. Full results across all models under global setting are provided in Supplementary Figs.~8–11. In the main text, we focus on LightGBM because it achieves the highest average R² and shows the most consistent performance across diverse urban contexts.


\subsection{Exploratory analysis of cross-city performance heterogeneity}\label{subsec4.5}
To interpret cross-city variation in predictive performance, we conducted an exploratory analysis focused on AlphaEarth, which showed the strongest overall performance across most domains and cities. First, we summarized each city’s predictive profile using mean test R² values for crime, income, health, and travel, and applied hierarchical clustering to identify groups of MSAs with similar cross-domain performance patterns. Second, we examined whether cross-city differences in AlphaEarth performance were descriptively associated with three urban form indicators: population density, employment and household entropy, and walkability. Metropolitan population totals were obtained from the U.S. Census Bureau Population Estimates Program\cite{BureauMetropolitan}, and population density was calculated as metropolitan population divided by MSA area. Employment and household entropy and walkability index were obtained from the EPA Smart Location Database\cite{USEPA2021Smart} and aggregated to the MSA scale. For each indicator, we fitted simple domain-specific linear trend lines and calculated Spearman’s rank correlation coefficients. These comparisons were intended as exploratory and descriptive, and were used to explore possible task-specific relationships between urban context and predictive performance.


\subsection{Dimensionality reduction methods}\label{subsec4.6}
To compare the high-dimensional latent representations of Prithvi (768-d) and Clay (1024-d) with the 64-d AlphaEarth embeddings on an equal footing, we applied five distinct dimensionality reduction techniques to compress the vectors to 64 dimensions.

Principal Component Analysis (PCA). A linear transformation that projects data onto the orthogonal directions of maximum variance\cite{2002Principal}.

Factor Analysis (FA). A statistical method that models observed variables as linear combinations of potential latent factors plus error variance\cite{Lawley1962Factor}.

 Kernel PCA (KPCA). A non-linear extension of PCA that uses kernel functions (specifically the Radial Basis Function) to project data into a higher-dimensional space where it is linearly separable before reduction\cite{Scholkopf1998Nonlinear}.
 
Isomap. A manifold learning technique that preserves the intrinsic geometric structure of the data by maintaining geodesic distances between points\cite{Tenenbaum2000Global}.

Random Projection. A computationally efficient method based on the Johnson-Lindenstrauss lemma, which preserves pairwise distances by projecting the data onto a lower-dimensional subspace using a random Gaussian matrix\cite{Bingham2001Random}.

All reduction methods were implemented using the scikit-learn framework in Python, with parameters fitted globally across all metropolitan areas to ensure consistent feature representation across spatial domains.

\section{Data availability}\label{sec5}
All datasets used in this study are publicly available. AlphaEarth embeddings were obtained through \href{https://earthengine.google.com/}{Google Earth Engine}. Prithvi embeddings were generated using the open-source Prithvi-EO-2.0 foundation model, available via \href{https://huggingface.co/ibm-nasa-geospatial/Prithvi-EO-2.0-300M}{Hugging Face} and \href{https://github.com/NASA-IMPACT/Prithvi-EO-2.0}{GitHub}. Clay embeddings were generated using the Clay foundation model (v1.5), available from the \href{https://clay-foundation.github.io/model/}{Clay Foundation Model} and \href{https://huggingface.co/made-with-clay}{Hugging Face}. Crime incident records for Houston, Los Angeles, New York City, Chicago, Atlanta and Seattle were obtained from the respective city open-data portals: \href{https://www.houstontx.gov/police/cs/Monthly_Crime_Data_by_Street_and_Police_Beat.htm}{Houston Police Department}, \href{https://data.lacity.org/Public-Safety/Crime-Data-from-2020-to-Present/2nrs-mtv8/about_data}{Los Angeles Police Department}, \href{https://data.cityofnewyork.us/Public-Safety/NYPD-Complaint-Data-Historic/qgea-i56i/about_data}{New York City Police Department}, \href{https://data.cityofchicago.org/Public-Safety/Crimes-2001-to-Present/ijzp-q8t2/about_data}{Chicago Police Department}, \href{https://opendata.atlantapd.org/}{Atlanta Police Department} and \href{https://data.seattle.gov/Public-Safety/SPD-Crime-Data-2008-Present/tazs-3rd5/about_data}{Seattle Police Department}. Income and travel indicators were derived from the ACS 5-year estimates, accessed via the \href{https://www.census.gov/programs-surveys/acs/data.html}{U.S. Census Bureau API}. Health indicators were obtained from the \href{https://data.cdc.gov/}{Centers for Disease Control and Prevention}. Metropolitan statistical area boundaries were obtained from the \href{https://www.census.gov/geographies/mapping-files/time-series/geo/tiger-line-file.html}{2020 TIGER/Line Shapefiles for Core Based Statistical Areas (CBSAs)} provided by the U.S. Census Bureau. Metropolitan population totals were obtained from the \href{https://www.census.gov/data/tables/time-series/demo/popest/2020s-total-metro-and-micro-statistical-areas.html}{U.S. Census Bureau metropolitan population estimates}. Employment and household entropy and walkability index data were obtained from the \href{https://www.epa.gov/smartgrowth/smart-location-mapping}{EPA Smart Location Database}.


\section{Acknowledgments}\label{sec7}
This material is partially based upon work supported by the National Science Foundation under 2430700, 2401860, and 2526487, and NASA under 80NSSC22KM0052. Any opinions, findings, and conclusions or recommendations expressed in this material are those of the author and the funders have no role in the study design, data collection, analysis, or preparation of this article.

\section{Supplementary information}\label{sec8}
Supplementary information is available for this paper.


\bibliography{sn-bibliography}

\end{document}


\vspace*{2cm}

\noindent{\Large\bfseries Supplementary Information}\\[0.5em]
\vspace{1.5em}

\noindent{\Large\bfseries Earth Embeddings Reveal Diverse Urban Signals from Space}\\[0.5em]

\vspace{1.5em}

\noindent
Wenjing Gong$^{1\dagger}$, 
Udbhav Srivastava$^{2\dagger}$, 
Yuchen Wang$^{1}$, 
Yuhao Jia$^{3}$, 
Qifan Wu$^{3}$, 
Weishan Bai$^{1}$, 
Yifan Yang$^{1}$, 
Xiao Huang$^{3*}$, 
Xinyue Ye$^{1,2*}$

\vspace{1.5em}

\noindent
$^{1}$Texas A\&M University, College Station, TX, USA. \\
$^{2}$The University of Alabama, Tuscaloosa, AL, USA. \\
$^{3}$Emory University, Atlanta, GA, USA.

\vspace{1em}

\noindent
*Corresponding author(s). E-mail(s): \textcolor{blue}{xiao.huang2@emory.edu}; \textcolor{blue}{xye10@ua.edu};

\noindent
$\dagger$These authors contributed equally to this work.

\newpage

\section*{List of Supplementary Items}

\noindent Supplementary Note 1: Details of the urban signals \dotfill 3

\noindent Supplementary Note 2: Mapping of Earth embeddings \dotfill 6

\noindent Supplementary Note 3: Performance comparison of supervised learning algorithms \dotfill 7

\noindent Supplementary Note 4: Model estimation results (LightGBM) \dotfill 9

\noindent Supplementary Note 5: Results for dimensionality-reduced embedding variants (LightGBM) \dotfill 11

\noindent Supplementary Figures 1--11 \dotfill 3--8

\noindent Supplementary Figures 12--13 \dotfill 9--10

\noindent Supplementary Table 1 \dotfill 3

\noindent Supplementary Table 2 \dotfill 9

\noindent Supplementary Tables 3--4 \dotfill 11--12


\newpage
\section{Supplementary Note 1: Details of the urban signals}\label{sec1}

\begin{table}[htbp]
\centering
\caption{Definitions of urban signals and their corresponding SDGs.}
\label{tab:sdg_indicator_definitions}
\begin{tabularx}{\textwidth}{
>{\raggedright\arraybackslash}p{0.26\textwidth}
>{\raggedright\arraybackslash}p{0.12\textwidth}
>{\raggedright\arraybackslash}p{0.16\textwidth}
X}
\toprule
\textbf{SDGs} & \textbf{Theme} & \textbf{Indicator} & \textbf{Definition} \\
\midrule

\multirow{2}{=}{Goal 11: Sustainable Cities and Communities}
& \multirow{2}{=}{Crime}
& Log (Violent Crime)
& Natural logarithm of the violent crime rate. \\
& & Log (Petty Crime)
& Natural logarithm of the petty crime rate. \\
\midrule

Goal 1: No Poverty
& Income
& Log (Median Income)
& Natural logarithm of median household income. \\
\midrule

\multirow{6}{=}{Goal 3: Good Health and Well-being}
& \multirow{6}{=}{Health}
& \%Obesity
& Model-based estimate for crude prevalence of obesity among adults. \\
& & \%Diabetes
& Model-based estimate for crude prevalence of diagnosed diabetes among adults. \\
& & \%Inactivity
& Model-based estimate for crude prevalence of no leisure-time physical activity among adults. \\
& & \%Mental Health
& Model-based estimate for crude prevalence of frequent mental distress among adults. \\
& & \%Physical Health
& Model-based estimate for crude prevalence of frequent physical distress among adults. \\
& & \%Cancer
& Model-based estimate for crude prevalence of cancer (non-skin) or melanoma among adults. \\
\midrule

\multirow{5}{=}{Goal 13: Climate Action}
& \multirow{5}{=}{Travel}
& \%Drive Alone
& Percentage of workers commuting by driving alone. \\
& & \%Transit
& Percentage of workers commuting by public transportation. \\
& & \%Bike
& Percentage of workers commuting by bicycle. \\
& & \%Walk
& Percentage of workers commuting by walking. \\
& & \%Bike+Walk
& Combined percentage of workers commuting by bicycle or walking. \\
\bottomrule
\end{tabularx}
\end{table}

\begin{figure}[H]
\centering
\includegraphics[width=0.9\textwidth]{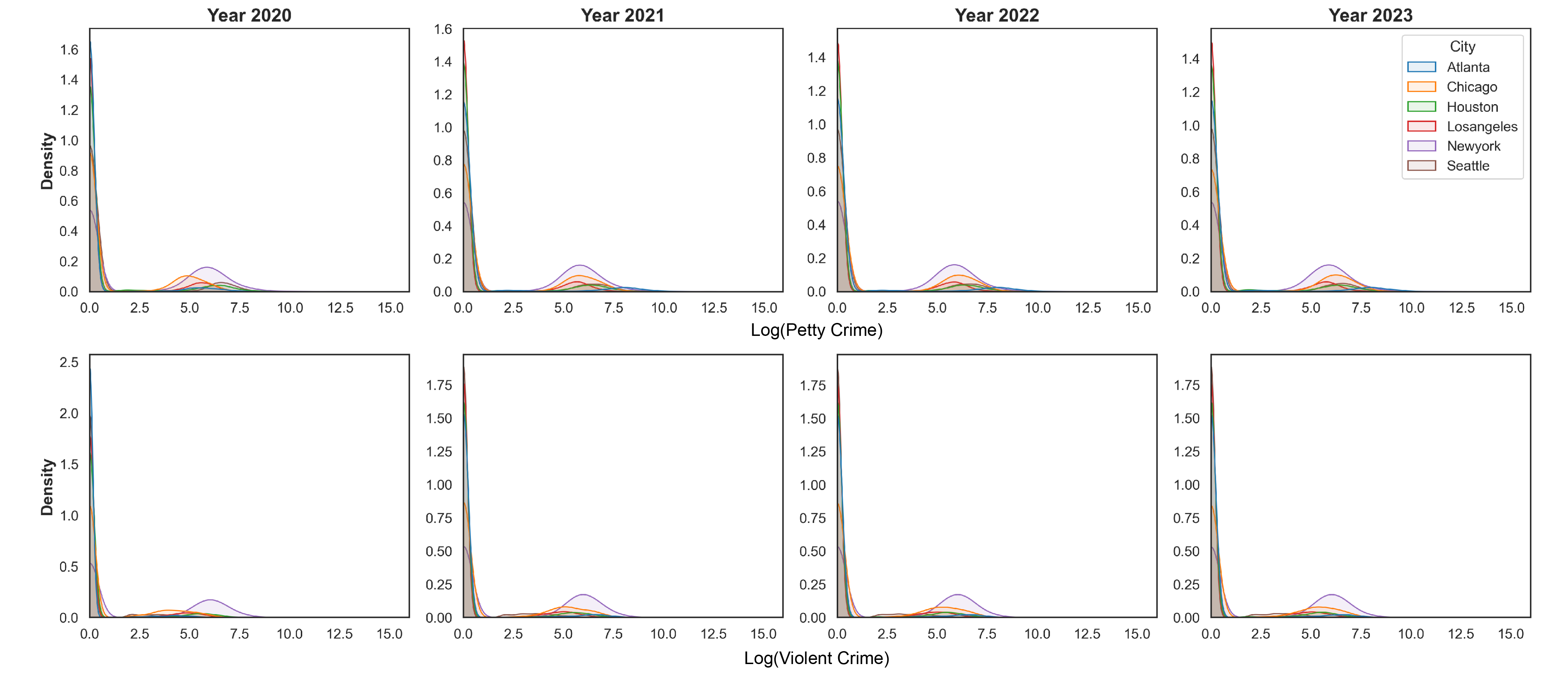}
\caption{Distribution of the crime indicators for six U.S. metropolitan areas, 2020–2023.}\label{fig12}
\end{figure}

\begin{figure}[H]
\centering
\includegraphics[width=0.9\textwidth]{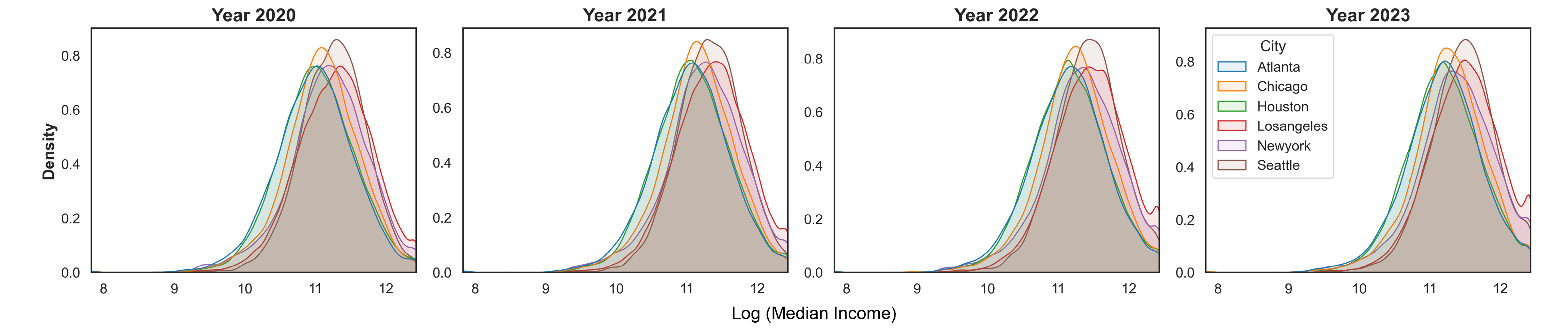}
\caption{Distribution of the income indicators for six U.S. metropolitan areas, 2020–2023.}\label{fig12}
\end{figure}

\begin{figure}[H]
\centering
\includegraphics[width=0.9\textwidth]{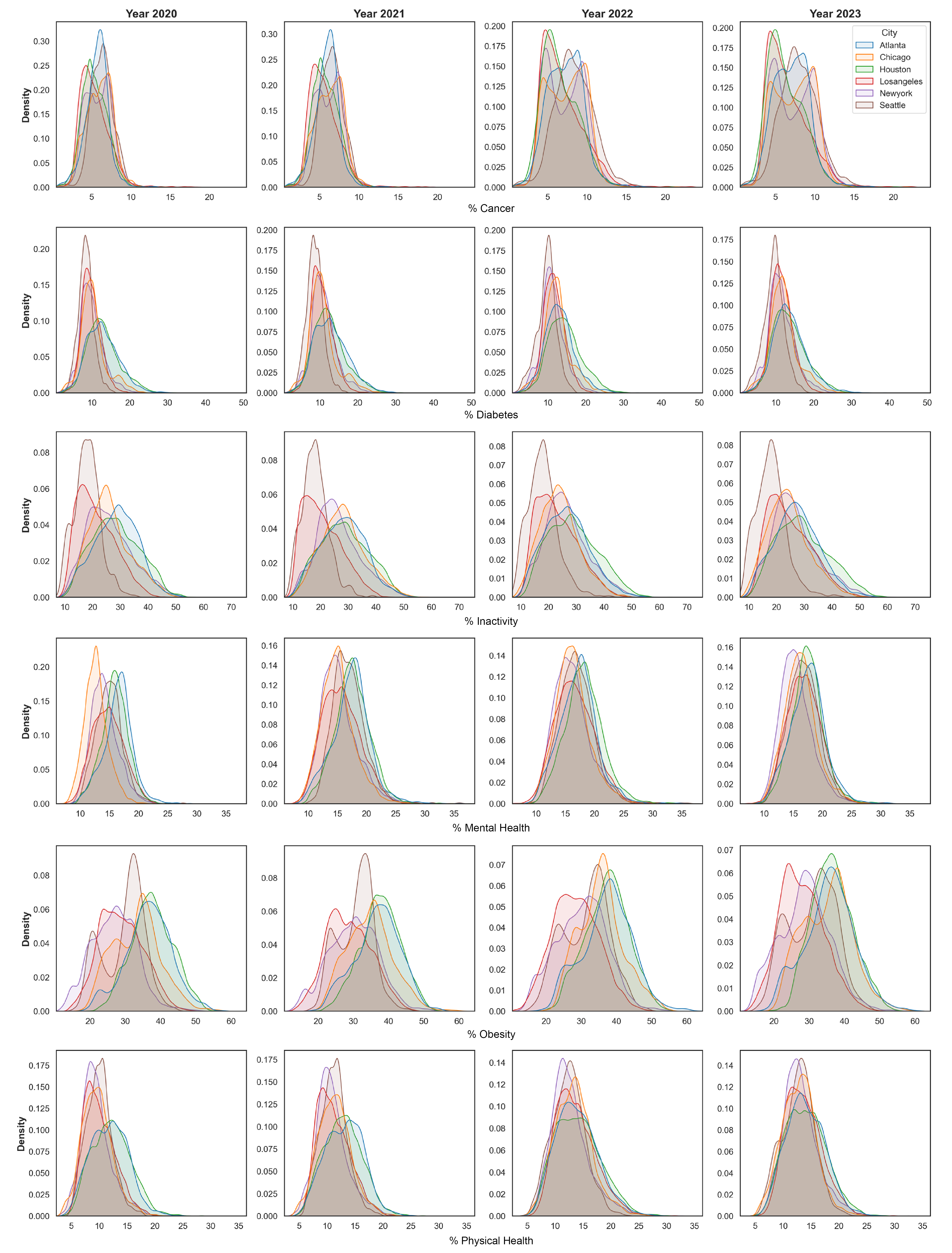}
\caption{Distribution of the health indicators for six U.S. metropolitan areas, 2020–2023.}\label{fig12}
\end{figure}

\begin{figure}[H]
\centering
\includegraphics[width=0.9\textwidth]{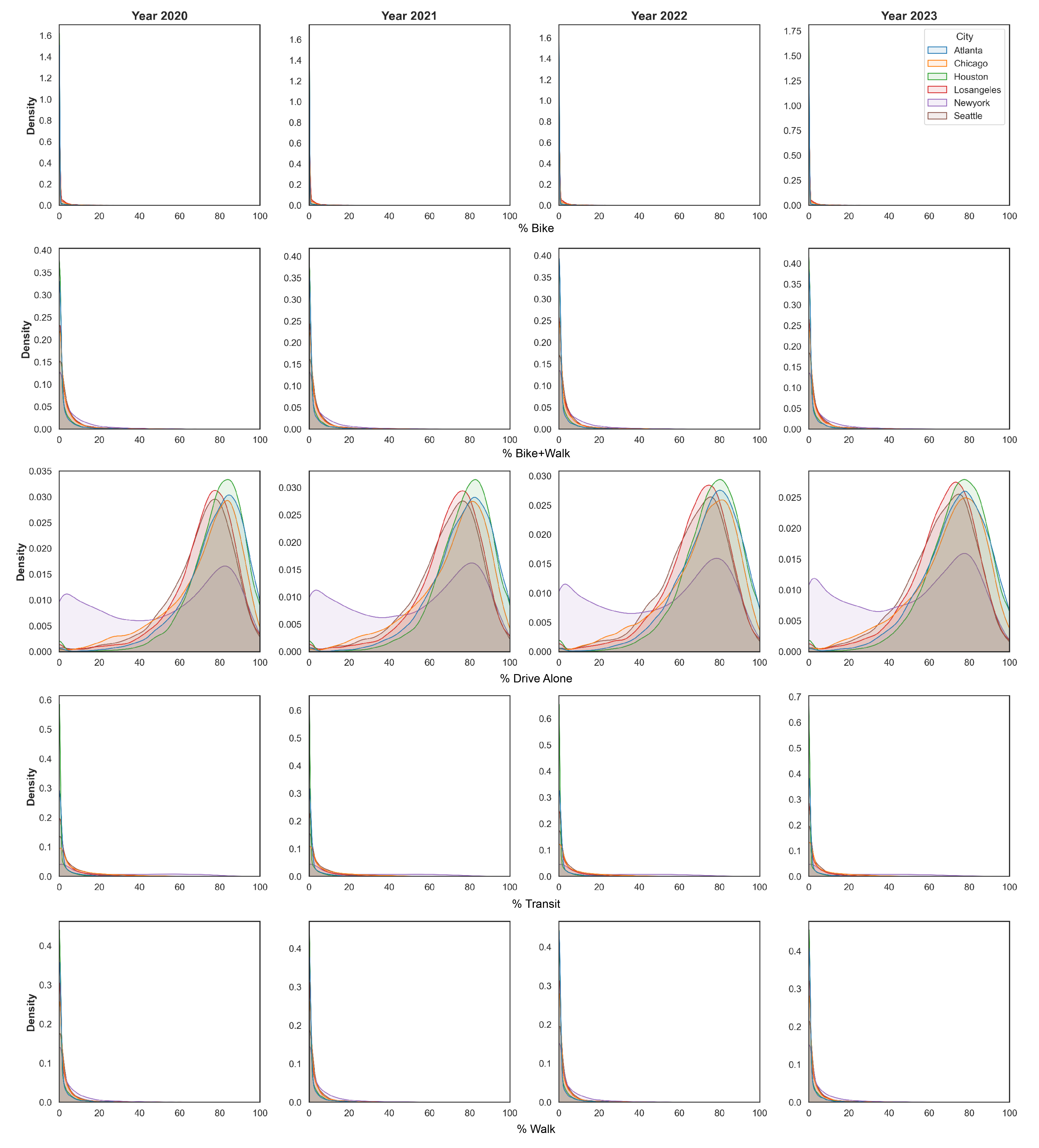}
\caption{Distribution of the travel indicators for six U.S. metropolitan areas, 2020–2023.}\label{fig12}
\end{figure}

\section{Supplementary Note 2: Mapping of Earth embeddings}\label{sec2}
\begin{figure}[H]
\centering
\includegraphics[width=0.8\textwidth]{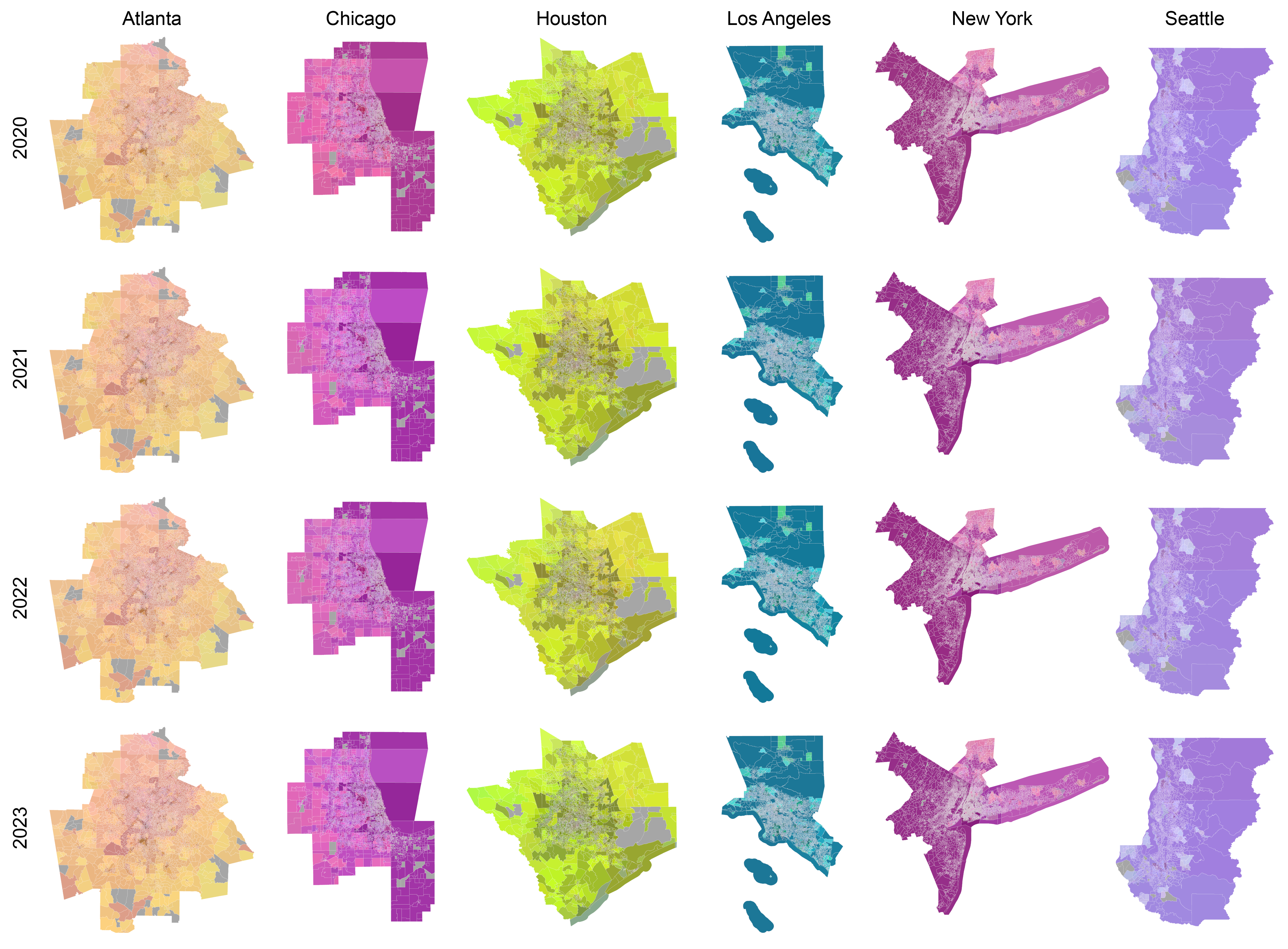}
\caption{AlphaEarth embedding maps for six U.S. metropolitan areas, aggregated to CBGs, 2020–2023. Colors represent a three-dimensional PCA projection of the embeddings, with PC1–PC3 mapped to the red, green, and blue channels. Hues are comparable across cities within a given year.}\label{fig12}
\end{figure}

\begin{figure}[H]
\centering
\includegraphics[width=0.8\textwidth]{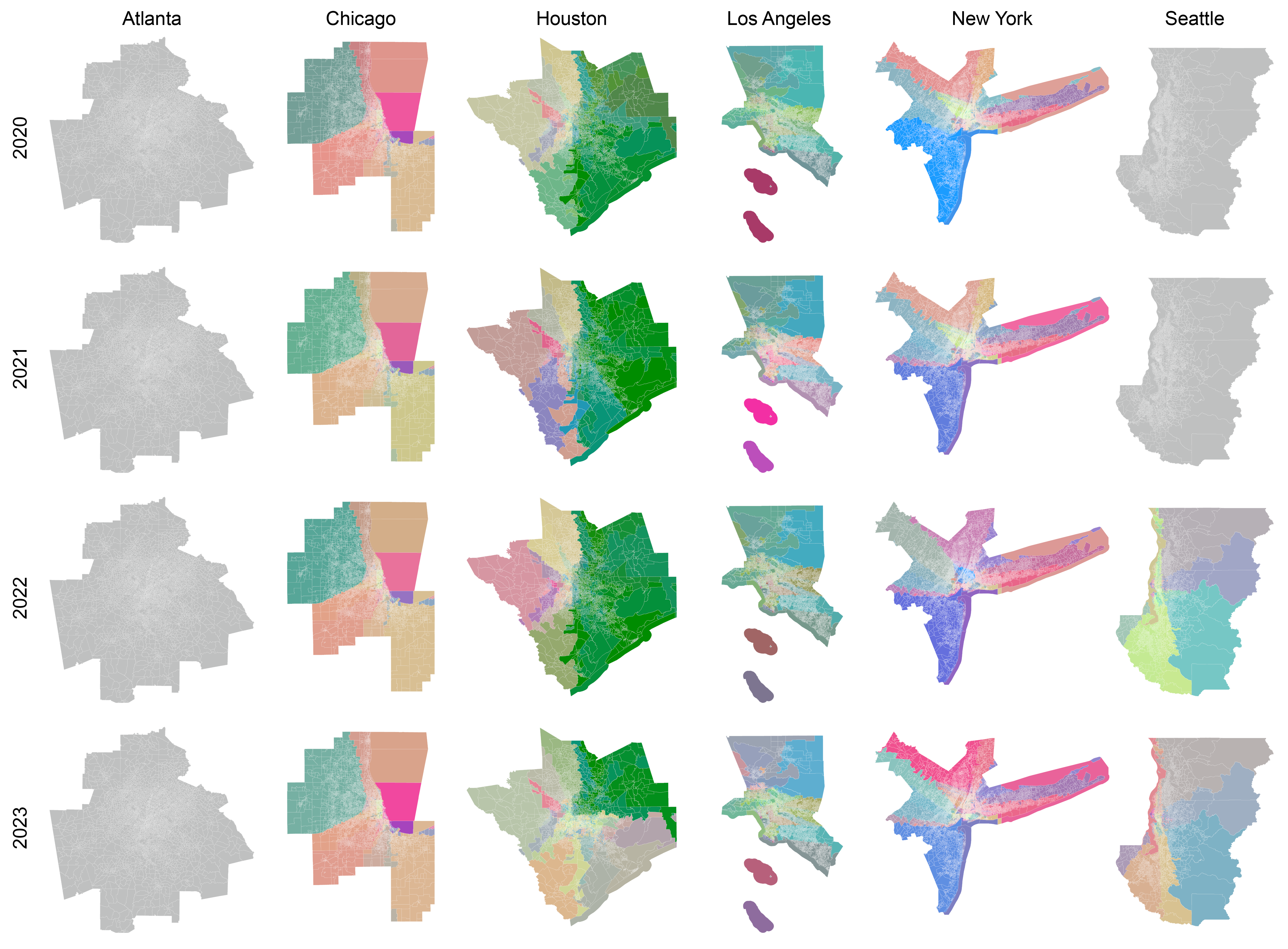}
\caption{Prithvi embedding maps for six U.S. metropolitan areas, aggregated to CBGs, 2020–2023. Colors represent a three-dimensional PCA projection of the embeddings, with PC1–PC3 mapped to the red, green, and blue channels. Hues are comparable across cities within a given year.}\label{fig12}
\end{figure}

\begin{figure}[H]
\centering
\includegraphics[width=0.8\textwidth]{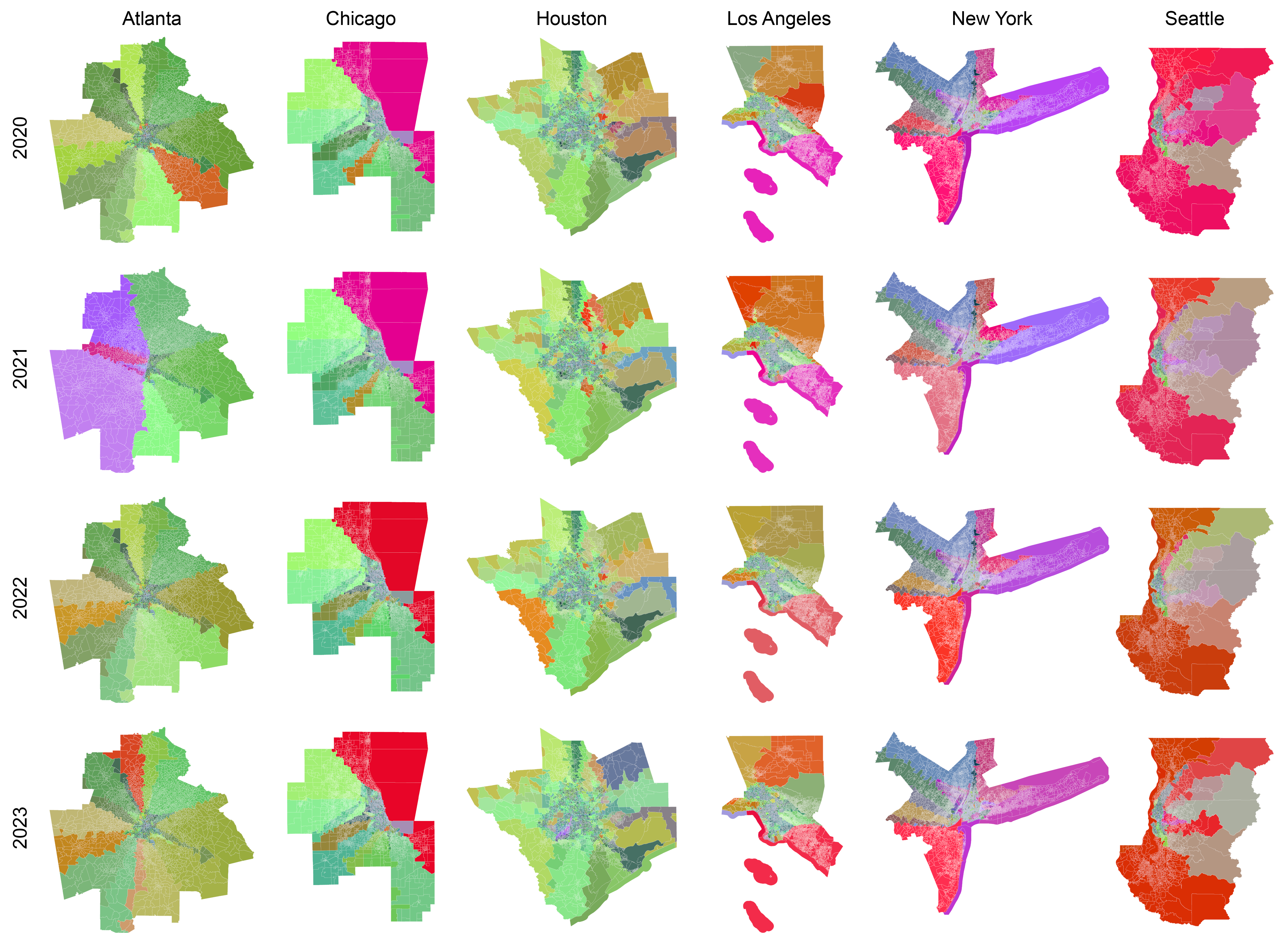}
\caption{Clay embedding maps for six U.S. metropolitan areas, aggregated to CBGs, 2020–2023. Colors represent a three-dimensional PCA projection of the embeddings, with PC1–PC3 mapped to the red, green, and blue channels. Hues are comparable across cities within a given year.}\label{fig12}
\end{figure}

\section{Supplementary Note 3: Performance comparison of supervised learning algorithms}\label{sec3}
\begin{figure}[H]
\centering
\includegraphics[width=0.9\textwidth]{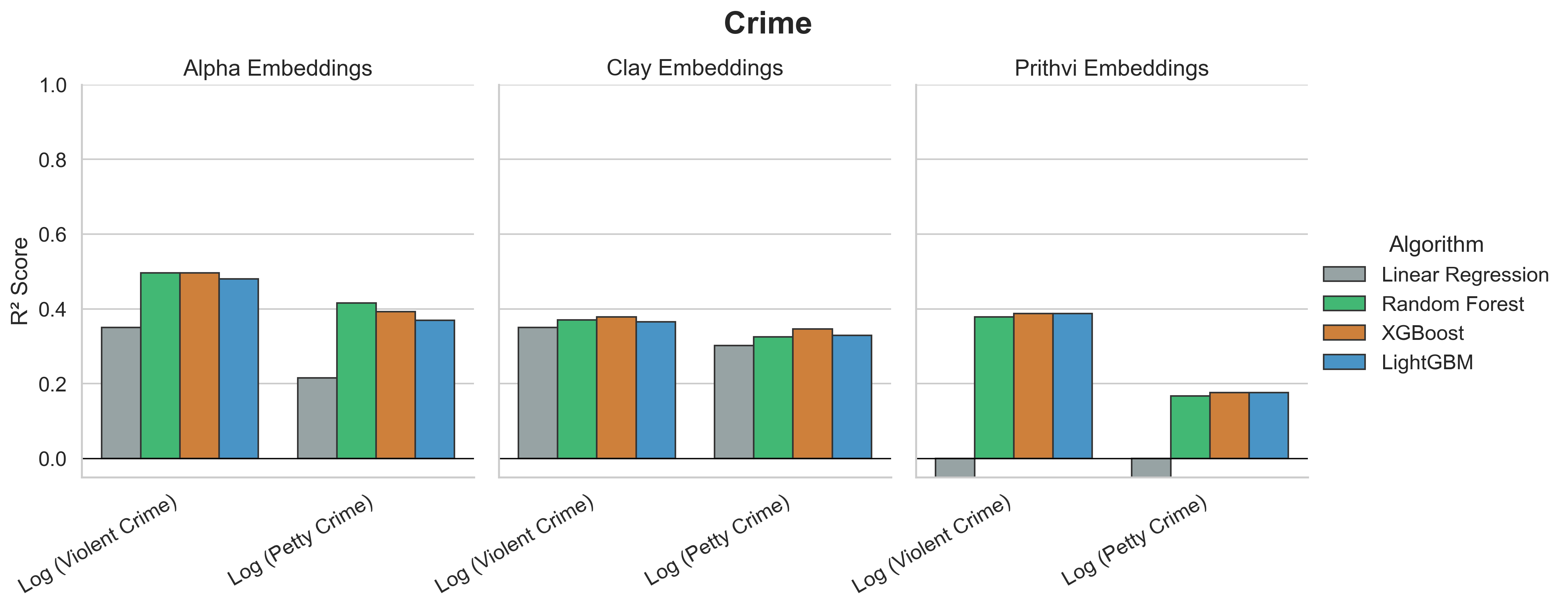}
\caption{Supervised ML models’ performance (test R²) of Earth observation embeddings for the crime theme (global setting).}\label{fig12}
\end{figure}

\begin{figure}[H]
\centering
\includegraphics[width=0.9\textwidth]{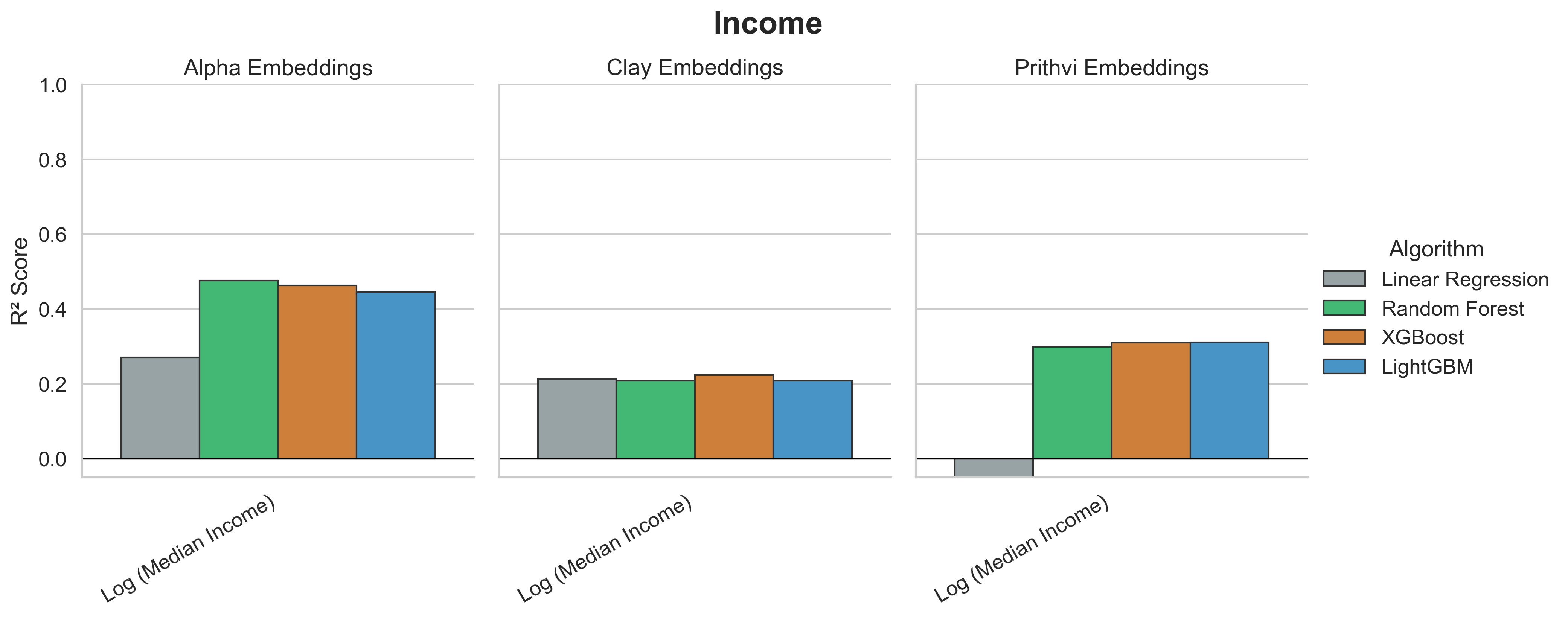}
\caption{Supervised ML models’ performance (test R²) of Earth embeddings for the income theme (global setting).}\label{fig12}
\end{figure}

\begin{figure}[H]
\centering
\includegraphics[width=0.9\textwidth]{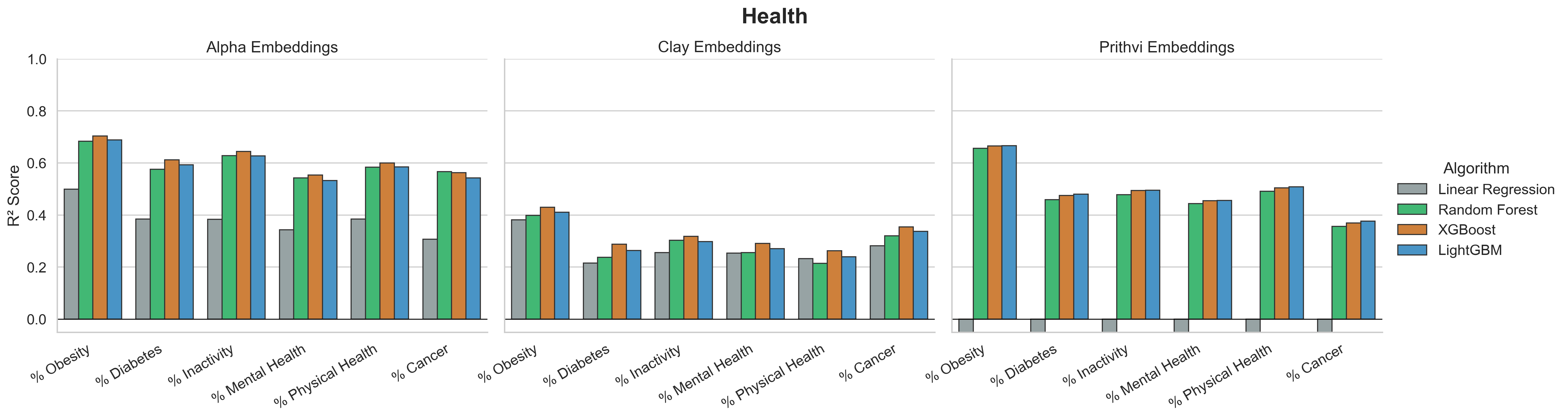}
\caption{Supervised ML models’ performance (test R²) of Earth embeddings for the health theme (global setting).}\label{fig12}
\end{figure}

\begin{figure}[H]
\centering
\includegraphics[width=0.9\textwidth]{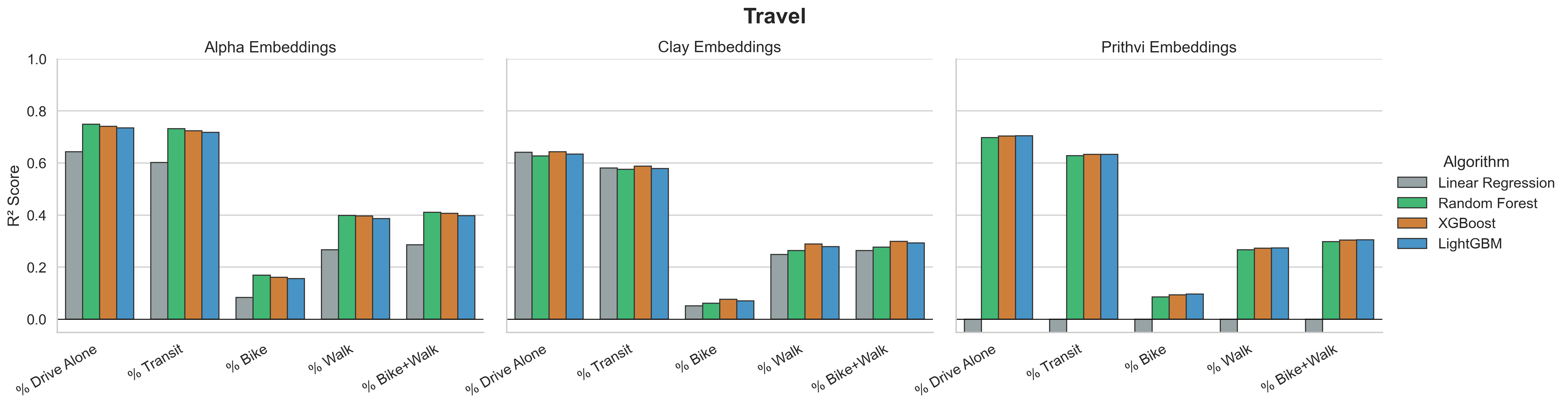}
\caption{Supervised ML models’ performance (test R²) of Earth embeddings for the travel theme (global setting).}\label{fig12}
\end{figure}

\section{Supplementary Note 4: Model estimation results (LightGBM)}\label{sec4}

\begin{table}[htbp]
\centering
\setlength{\tabcolsep}{4pt}
\caption{Indicator-level test $R^2$ of Earth embeddings across all MSAs and years (global setting).}

\begin{tabular*}{\textwidth}{@{\extracolsep{\fill}}l p{4cm} ccc}
\toprule
Theme & Indicator & AlphaEarth & Prithvi & Clay \\
\midrule

\multirow{2}{*}{Crime}
& Log (Violent) & \textbf{0.480} & 0.388 & 0.366 \\
& Log (Petty)   & \textbf{0.369} & 0.176 & 0.329 \\

\midrule

Income
& Log (Median Income) & \textbf{0.444} & 0.311 & 0.209 \\

\midrule

\multirow{6}{*}{Health}
& \%Obesity          & \textbf{0.689} & 0.666 & 0.411 \\
& \%Diabetes         & \textbf{0.593} & 0.480 & 0.264 \\
& \%Inactivity       & \textbf{0.627} & 0.495 & 0.298 \\
& \%Mental Health    & \textbf{0.532} & 0.456 & 0.270 \\
& \%Physical Health  & \textbf{0.585} & 0.509 & 0.240 \\
& \%Cancer           & \textbf{0.543} & 0.377 & 0.337 \\

\midrule

\multirow{5}{*}{Travel}
& \%Drove Alone & \textbf{0.735} & 0.705 & 0.634 \\
& \%Transit     & \textbf{0.717} & 0.633 & 0.578 \\
& \%Bike        & \textbf{0.156} & 0.096 & 0.071 \\
& \%Walk        & \textbf{0.387} & 0.274 & 0.278 \\
& \%Bike + Walk & \textbf{0.398} & 0.305 & 0.293 \\

\bottomrule
\end{tabular*}

\medskip
\footnotesize
\parbox{\textwidth}{
Note: Results are reported at the CBG level, except for indicators in the health theme, which are only available at the CT level. Boldface indicates the best-performing model for each indicator.
}

\end{table}

\begin{figure}[H]
\centering
\includegraphics[width=0.9\textwidth]{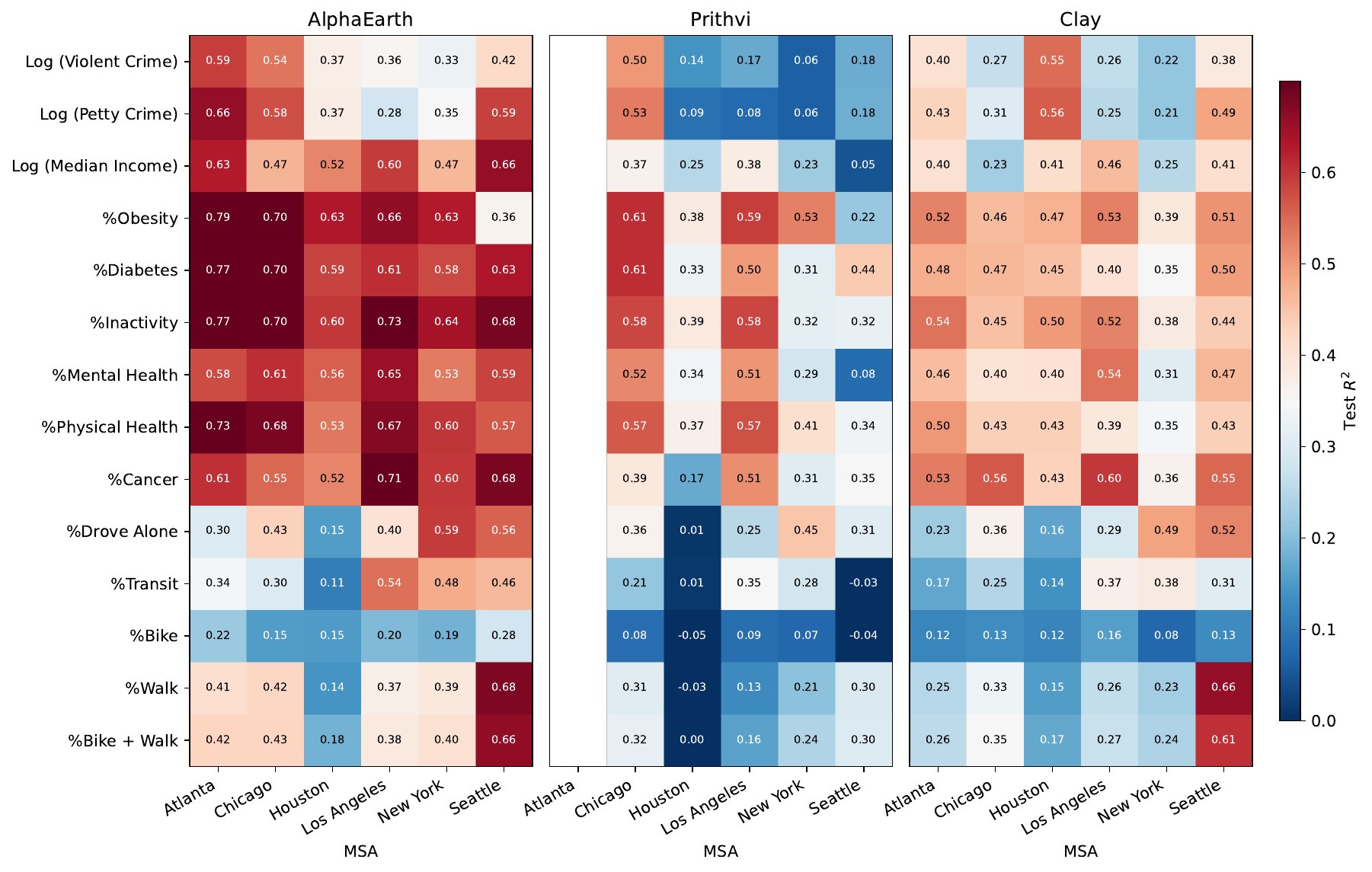}
\caption{City-wise benchmark performance (test R²) of Earth embeddings across all years (2020–2023).}\label{fig12}
\end{figure}

\begin{figure}[H]
\centering
\includegraphics[width=0.9\textwidth]{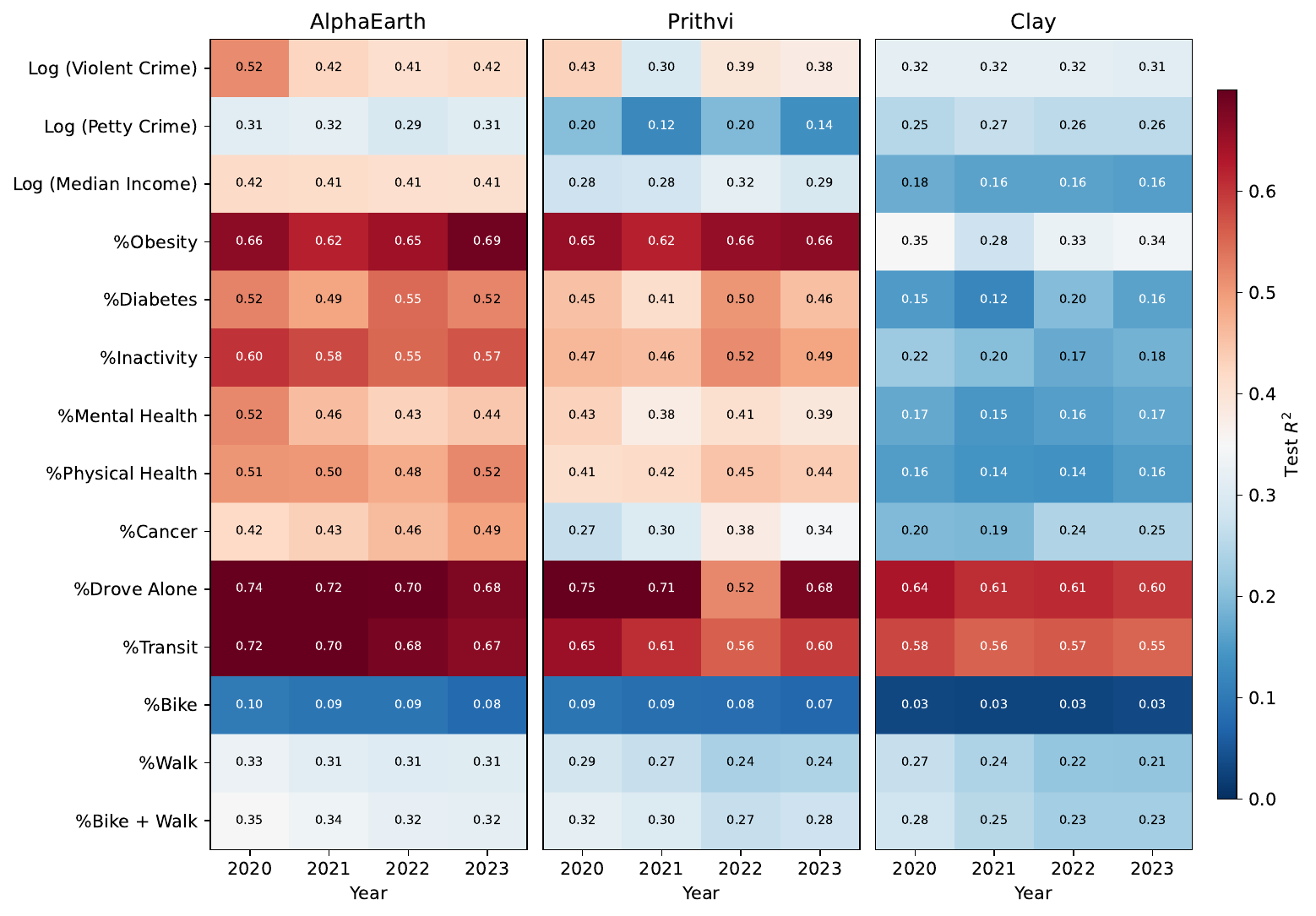}
\caption{Year-wise benchmark performance (test R²) of Earth embeddings across all six MSAs.}\label{fig13}
\end{figure}

\newpage

\section{Supplementary Note 5: Results for dimensionality-reduced embedding variants}\label{sec5}

\begin{table}[htbp]
\centering
\small
\setlength{\tabcolsep}{3pt}
\caption{Indicator-level test $R^2$ of Prithvi embeddings and their dimensionality-reduced variants across all MSAs and years (global setting).}

\begin{tabular}{l p{2cm} ccccccc}
\toprule
Theme & Indicator & \cellcolor{gray!25}AlphaEarth & Prithvi & Prithvi\_fa & Prithvi\_isomap & Prithvi\_kpca & Prithvi\_pca & Prithvi\_rp \\
\midrule

\multirow{2}{*}{Crime}
& Log (Violent Crime) & \cellcolor{gray!25}0.480 & \textbf{0.388} & 0.385 & 0.349 & 0.367 & 0.385 & 0.373 \\
& Log (Petty Crime)   & \cellcolor{gray!25}0.369 & \textbf{0.176} & 0.164 & 0.153 & 0.161 & 0.167 & 0.167 \\

\midrule

Income & Log (Median Income) & \cellcolor{gray!25}0.444 & \textbf{0.311} & 0.299 & 0.278 & 0.290 & 0.300 & 0.301 \\

\midrule

\multirow{6}{*}{Health}
& \%Obesity          & \cellcolor{gray!25}0.689 & \textbf{0.666} & 0.620 & 0.541 & 0.584 & 0.615 & 0.636 \\
& \%Diabetes         & \cellcolor{gray!25}0.593 & \textbf{0.480} & 0.456 & 0.383 & 0.419 & 0.443 & 0.461 \\
& \%Inactivity       & \cellcolor{gray!25}0.627 & \textbf{0.495} & 0.470 & 0.397 & 0.429 & 0.464 & 0.473 \\
& \%Mental Health    & \cellcolor{gray!25}0.532 & \textbf{0.456} & 0.432 & 0.361 & 0.407 & 0.435 & 0.435 \\
& \%Physical Health  & \cellcolor{gray!25}0.585 & \textbf{0.509} & 0.488 & 0.409 & 0.458 & 0.483 & 0.484 \\
& \%Cancer           & \cellcolor{gray!25}0.543 & \textbf{0.377} & 0.346 & 0.278 & 0.314 & 0.349 & 0.351 \\

\midrule

\multirow{5}{*}{Travel}
& \%Drive Alone & \cellcolor{gray!25}0.735 & \textbf{0.705} & 0.696 & 0.668 & 0.687 & 0.700 & 0.692 \\
& \%Transit     & \cellcolor{gray!25}0.717 & \textbf{0.633} & 0.630 & 0.613 & 0.626 & 0.631 & 0.625 \\
& \%Bike        & \cellcolor{gray!25}0.156 & 0.096 & \textbf{0.098} & 0.091 & 0.094 & 0.095 & 0.097 \\
& \%Walk        & \cellcolor{gray!25}0.387 & 0.274 & 0.264 & 0.276 & \textbf{0.278} & 0.266 & 0.272 \\
& \%Bike + Walk & \cellcolor{gray!25}0.398 & 0.305 & 0.295 & 0.304 & \textbf{0.306} & 0.297 & 0.304 \\

\bottomrule
\end{tabular}

\medskip
\footnotesize
\parbox{\textwidth}{Note: Results are reported at the CBG level, except for indicators in the health theme, which are only available at the CT level. Boldface indicates the best-performing model for each indicator; grey shading denotes the AlphaEarth baseline results.}
\end{table}


\begin{table}[htbp]
\centering
\small
\setlength{\tabcolsep}{3pt}
\caption{Indicator-level test $R^2$ of Clay embeddings and their dimensionality-reduced variants across all MSAs and years (global setting).}

\begin{tabular}{l p{2cm} ccccccc}
\toprule
Theme & Indicator & \cellcolor{gray!25}AlphaEarth & Clay & Clay\_fa & Clay\_isomap & Clay\_kpca & Clay\_pca & Clay\_rp \\
\midrule

\multirow{2}{*}{Crime}
& Log (Violent Crime) & \cellcolor{gray!25}0.480 & \textbf{0.366} & 0.166 & 0.133 & 0.171 & 0.162 & 0.324 \\
& Log (Petty Crime)   & \cellcolor{gray!25}0.369 & \textbf{0.329} & 0.119 & 0.096 & 0.127 & 0.117 & 0.257 \\

\midrule

Income & Log (Median Income) & \cellcolor{gray!25}0.444 & \textbf{0.208} & 0.087 & 0.081 & 0.091 & 0.087 & 0.171 \\

\midrule

\multirow{6}{*}{Health}
& \%Obesity          & \cellcolor{gray!25}0.689 & \textbf{0.411} & 0.300 & 0.112 & 0.156 & 0.119 & 0.292 \\
& \%Diabetes         & \cellcolor{gray!25}0.593 & \textbf{0.263} & 0.097 & 0.046 & 0.068 & 0.053 & 0.187 \\
& \%Inactivity       & \cellcolor{gray!25}0.627 & \textbf{0.298} & 0.142 & 0.074 & 0.102 & 0.070 & 0.221 \\
& \%Mental Health    & \cellcolor{gray!25}0.532 & \textbf{0.270} & 0.153 & 0.083 & 0.096 & 0.087 & 0.213 \\
& \%Physical Health  & \cellcolor{gray!25}0.585 & \textbf{0.240} & 0.115 & 0.075 & 0.100 & 0.082 & 0.165 \\
& \%Cancer           & \cellcolor{gray!25}0.543 & \textbf{0.337} & 0.173 & 0.122 & 0.149 & 0.116 & 0.301 \\

\midrule

\multirow{5}{*}{Travel}
& \%Drive Alone & \cellcolor{gray!25}0.735 & \textbf{0.634} & 0.366 & 0.294 & 0.334 & 0.355 & 0.586 \\
& \%Transit     & \cellcolor{gray!25}0.717 & \textbf{0.578} & 0.372 & 0.317 & 0.325 & 0.355 & 0.562 \\
& \%Bike        & \cellcolor{gray!25}0.156 & \textbf{0.071} & 0.020 & 0.016 & 0.019 & 0.020 & 0.055 \\
& \%Walk        & \cellcolor{gray!25}0.387 & \textbf{0.278} & 0.129 & 0.113 & 0.128 & 0.129 & 0.239 \\
& \%Bike + Walk & \cellcolor{gray!25}0.398 & \textbf{0.293} & 0.136 & 0.118 & 0.135 & 0.138 & 0.251 \\

\bottomrule
\end{tabular}

\medskip
\footnotesize
\parbox{0.9\textwidth}{Note: Results are reported at the CBG level, except for indicators in the health theme, which are only available at the CT level. Boldface indicates the best-performing model for each indicator; grey shading denotes the AlphaEarth baseline results.}

\end{table}
